\documentclass[twocolumn]{IEEEtran}

\input{mysymbol.sty}
\usepackage{graphicx,dsfont}
\usepackage{epsfig}
\usepackage{epstopdf}
\graphicspath{{./eps/}}
\usepackage{amsmath}
\usepackage{bbm}
\usepackage{amssymb}
\usepackage{wrapfig}
\usepackage{times,color}
\usepackage{bbold}
\usepackage{cite,footnote,xspace,syntonly,algorithm,algorithmic,bm}
\usepackage{algorithm,algorithmic}
\usepackage{mdwmath}
\usepackage{mdwtab}
\usepackage{eqparbox}

\usepackage{placeins}
\usepackage{float}
\usepackage{tabularx,colortbl}
\usepackage{url}
\usepackage{caption}
\usepackage{subcaption}

\def \A {{\mathbf{A}}}
\def \a {{\mathbf{a}}}

\def \I {{\mathbf{I}}}

\def \S {{\mathbf{S}}}
\def \X {{\mathbf{X}}}
\def \E {{\mathbf{E}}}
\def \e {{\mathbf{e}}}
\def \O {{\mathbf{O}}}
\def \o {{\mathbf{o}}}
\def \U {{\mathbf{U}}}
\def \V {{\mathbf{V}}}
\def \P {{\mathbf{P}}}
\def \W {{\mathbf{W}}}
\def \Z {{\mathbf{Z}}}
\def \u {{\mathbf{u}}}
\def \v {{\mathbf{v}}}

\long\def\symbolfootnote[#1]#2{\begingroup
\def\thefootnote{\fnsymbol{footnote}}
\footnote[#1]{#2}\endgroup} \psfull

\begin{document}


\title{\huge Joint Community and Anomaly Tracking \\in Dynamic Networks$^\dag$}

\author{{\it Brian Baingana, \textit{Student Member}, \textit{IEEE} and Georgios~B.~Giannakis, \textit{Fellow}, \textit{IEEE}$^\ast$}}

\markboth{IEEE TRANSACTIONS OF SIGNAL PROCESSING (SUBMITTED)}
\maketitle \maketitle \symbolfootnote[0]{$\dag$ Work in this paper was 
supported by NSF-EECS Grant No. 1343248 and AFOSR-MURI Grant No. 
FA9550-10-1-0567.
Parts of the paper appeared in the {\it Proc. of
the IEEE Global Conference on Signal and Information Processing}, Atlanta, Georgia, December 3-5, 2014.} 
\symbolfootnote[0]{$\ast$ The authors are with the Dept.
of ECE and the Digital Technology Center, University of
Minnesota, 200 Union Street SE, Minneapolis, MN 55455. Tel/fax:
(612)626-7781/625-4583; Emails:
\texttt{\{baing011,georgios\}@umn.edu}}



\thispagestyle{empty}\addtocounter{page}{-1}
\begin{abstract}
Most real-world networks exhibit community structure, a phenomenon
characterized by existence of node clusters whose intra-edge connectivity
is stronger than edge connectivities between nodes belonging to 
different clusters. In addition to facilitating a better understanding of
network behavior, community detection finds many practical applications 
in diverse settings. Communities in online social
networks are indicative of shared functional roles, or affiliation
to a common socio-economic status, the knowledge of which is vital for
targeted advertisement. In buyer-seller networks, community detection 
facilitates better product recommendations. Unfortunately, reliability of
community assignments is hindered by anomalous user behavior often 
observed as unfair self-promotion, or ``fake'' highly-connected
accounts created to promote fraud. The present paper advocates a 
novel approach for jointly tracking communities while detecting such 
anomalous nodes in time-varying networks. By postulating edge
creation as the result of mutual community participation by node pairs,
a dynamic factor model with anomalous memberships captured through 
a sparse outlier matrix is put forth. Efficient tracking algorithms
suitable for both online and decentralized operation are developed.
Experiments conducted on both synthetic and real network time series
successfully unveil underlying communities and anomalous nodes.
\end{abstract}


%


\section{Introduction}
\label{sec:introduction}
Networks underlie many complex phenomena involving
pairwise interactions between entities~\cite{kleinberg_book,kolaczyk_book}.
Examples include online social networks such as Facebook or Twitter,
e-mail and phone correspondences among individuals,
the Internet, and electric power grids. Most network analyses
focus on static networks, with node and link structures
assumed fixed. However, real-world networks often
evolve over time e.g., new links are frequently added to the web. 
Incorporating such temporal dynamics
plays a fundamental role towards a better understanding of network behavior.  

Community identification is one of the most 
studied tasks in modern network analysis
~\cite{fortunato2010community,girvan}. Fundamentally,
communities pertain to the inherent grouping of nodes, 
with many edges connecting nodes belonging to the same cluster,
and far fewer edges existing between clusters.
Cognizant of the temporal behavior inherent to networks,
several recent works have focused on the task of tracking time-varying communities
~\cite{xu,mankad,tang2008community,lin2009,greene2010tracking}.
Identification of dynamic communities finds applications in many 
settings e.g., grouping subscribers of 
an online social network into functional roles for more informative
advertising, or clustering blogs into content groups,
facilitating improved recommendations to readers.

Classical community detection 
approaches predominantly resort to well-studied unsupervised machine learning
algorithms e.g., hierarchical 
and spectral clustering~\cite{hastie_book,johnson1967hierarchical,von2007tutorial}.
To facilitate interpretability, several authors
have postulated that a network exhibits community
structure if it contains subnetworks whose expected node degree 
exceeds that of a random graph; see e.g., modularity~\cite{newman2006modularity}.
Many of these methods conduct hard community assignment,
whereby no node can jointly belong to more than one community. 
Nevertheless, communities in real networks tend to overlap with, 
or even completely contain others e.g., a Facebook user 
may jointly belong to a circle of college friends, 
and another comprising workplace colleagues. The quest to unveil possibly
overlapping communities has been at the forefront of
efforts to develop more flexible community discovery algorithms, capable of
associating each node with a per-community affiliation strength a.k.a,
soft clustering. Among these, factor models e.g.,
non-negative matrix factorization (NMF), 
have recently become popular for overlapping community discovery~\cite{mankad,rossi1,yang2013overlapping}.

\begin{figure}[t]
\centering
\begin{subfigure}[t]{.24\textwidth}
  \centering
  \includegraphics[width=3.5cm]{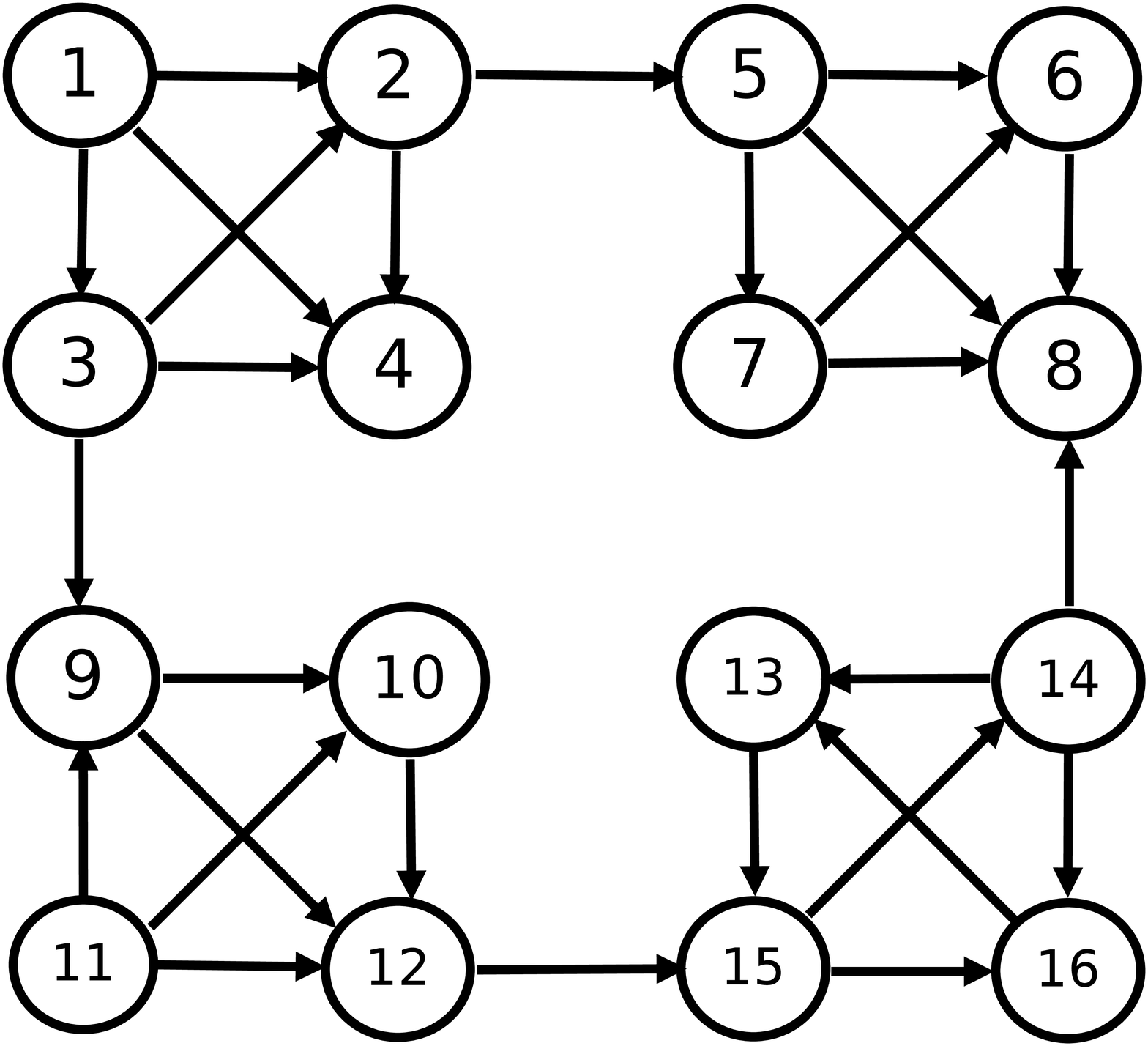}
  \caption{ } 
\end{subfigure}
\begin{subfigure}[t]{0.24\textwidth}
  \centering
 \includegraphics[width=3.5cm]{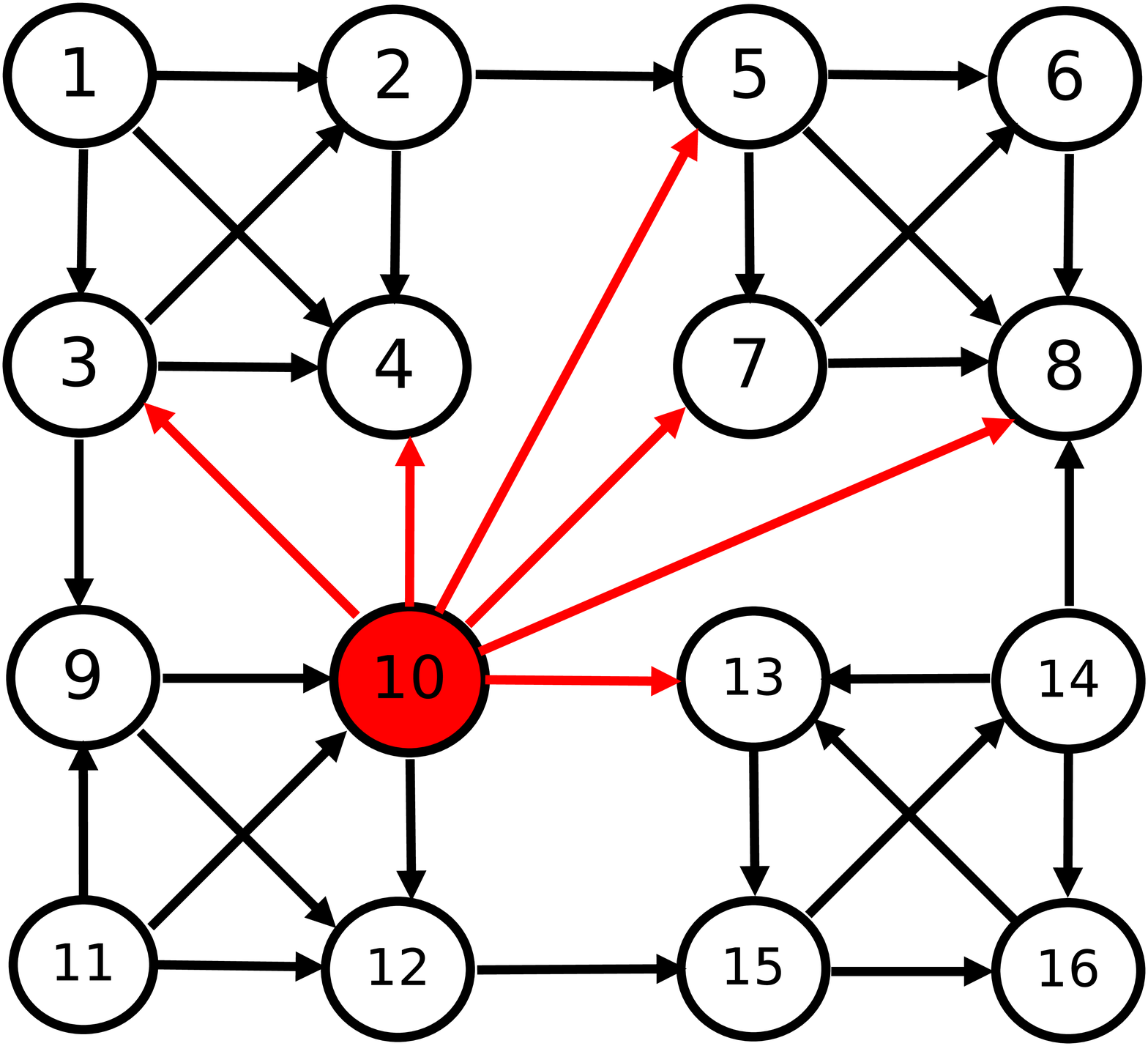}
  \caption{ } 
\end{subfigure}
\caption{ An illustration of the distortive effect of anomalous
nodes: a) A $16$-node directed network with four easily-discernible communities; and 
b) The same network with node $10$ exhibiting an unusually high number
of outgoing edges. Identifying the underlying communities is more challenging,
as a result of the distortion caused by node $10$.}
\label{fig:anets}
\end{figure}

This paper builds upon recent advances in overlapping 
community identification, with focus on dynamic networks. It is
assumed that temporal variations are slow across observation time
intervals. In addition, special attention is paid to 
existence of aberrant nodes exhibiting ``anomalous'' behavior.
Such behavior may often manifest as unusually strong, uni-directional 
edge connectivity across communities, leading to distortion 
of true communities; see Figure~\ref{fig:anets}. 
Examples include e-mail spammers, or individuals with 
malicious intent, masquerading under ``false'' Facebook profiles 
to initiate to connect with as many
legitimate users as possible. Anomalies identification
facilitates discovery of more realistic communities.
The present paper develops algorithms for 
jointly tracking time-varying communities,
while compensating for anomalies.

Several prior works have studied
the evolution of general temporal behavior in time-varying networks; see e.g.,~\cite{asur,backstrom,dunlavy,fu2009dynamic,sun2007graphscope,papadimitriou2005streaming}. 
Using tensor and matrix factorizations, 
temporal link prediction approaches are advocated for bipartite graphs
in~\cite{dunlavy}. Postulating a state-space 
dynamic stochastic blockmodel for dynamic edge evolution,
an extended Kalman filter was proposed for tracking communities
in~\cite{xu}. An NMF model was advocated
for batch recovery of overlapping communities
in time-varying networks in~\cite{mankad}. Generally, 
these contemporary approaches do not account for the 
occurrence of distortive anomalies, 
which may hurt the accuracy of community assignments.

The fresh look advocated by the present paper jointly accounts
for temporal variations and anomalous affiliations. Motivated by contemporary 
NMF approaches, a community affiliation model in which edge weights are 
generated by mutual community participation between node pairs is adopted
in Section~\ref{sec:model}. 
The proposed approach capitalizes on sparsity of anomalies, rank deficiency inherent to
networks with far fewer communities than the network size, as well as
slow edge variation across time intervals.
Under these conditions, a sparsity-promoting, rank-regularized, 
exponentially-weighted least-squares estimator is put forth in 
Section~\ref{sec:ctalg}.
Leveraging advances in proximal splitting approaches (see e.g.,~\cite{beck}), 
computationally efficient community tracking algorithms,
based on alternating minimization are developed. 

In order to appeal to big data contexts, within which most social
networks of interest arise, a number of algorithmic modifications
are considered. Towards facilitating real-time, memory-efficient
operation in streaming data settings, an online
algorithm leveraging stochastic gradient descent iterations is developed in Section~\ref{ssec:sgd}. Moreover, certain practical settings entail
storage of network data across many clusters of computing nodes, possibly geographically located at different sites. In such scenarios, tracking communities
in a decentralized fashion is well-motivated. 
To this end, a tracking algorithm that leverages
the alternating direction method of multipliers (ADMM) is developed
in Section~\ref{sec:decent}. Numerical tests with synthetically generated networks demonstrate
the effectiveness of the developed algorithms in tracking
communities and anomalies (Section~\ref{ssec:synthdata}). 
Further experiments in Section~\ref{ssec:realdat} are conducted
on real data, extracted from global trade flows among nations
between $1870$ and $2009$. 

\noindent\textit{Notation}. Bold uppercase (lowercase) letters will denote matrices
(column vectors), while operators $(\cdot)^{\top}$, $\lambda_{\max}(\cdot)$, and
$\textrm{diag}(\cdot)$ will stand for matrix transposition, maximum eigenvalue, 
and diagonal matrix, respectively. The identity matrix will be represented by $\I$, while $\mathbf{0}$ denotes the matrix of all zeros. The $\ell_p$ and Frobenius norms will be denoted by $\|\cdot\|_p$, and $\|\cdot\|_F$, respectively. The indicator function 
$\mathbb{1}_{\{x\}} = 1$ if 
$x$ evaluates to ``true'', otherwise $\mathbb{1}_{\{x\}} = 0$. Finally, $[\X]_{+}$ will
denote projection of $\X$ onto the non-negative orthant, with 
the $(i,j)$-th entry $\big[ [\mathbf{X}]_{+} \big]_{ij} = \text{max}([\X]_{ij},0)$.


\section{Model and Problem Statement}
\label{sec:model}
\subsection{Community affiliation model}
\label{ssec:camodel}
Consider a dynamic directed $N$-node network whose time-varying topology is captured
by the time-series of adjacency matrices 
$ \{ \A^t \in \mathbb{R}^{N \times N} \}_{t=1}^T$.
Entry $(i,j)$ of $\A^t$ (hereafter denoted by
$a_{ij}^t$) is nonzero only if an edge originating from
node $i$ is connected to node $j$ during interval $t$. 
It is assumed that edge weights are
nonnegative, namely $a_{ij}^t \geq 0$. Suppose that the network consists
of $C$ unknown communities which are allowed 
to overlap, that is a node can belong to
one or more communities simultaneously. This is motivated
by practical settings where e.g., a Facebook user
may be associated with multiple communities
consisting of her work colleagues, former schoolmates, or
friends from the local sports club. It can be reasoned
that the likelihood of two people becoming friends 
is directly related to the number of communities to
which they mutually belong. For example, if two work 
colleagues happen to have attended the same high school,
then chances are high that they will become friends.
This a fortiori argument based upon a reasonably
natural observation in social settings lies at the
foundation for several recent community affiliation 
models for edge generation~\cite{yang2013overlapping,mankad}.

Suppose $\V^t := [\v_1^t \dots \v_C^t] \in \mathbb{R}^{N \times C}$ denotes a temporal
basis matrix whose columns span a linear subspace of 
dimension $C$ during observation interval $t$. Associating
each basis vector with one of $C$ communities, the edge vector
associated with each node $i$ can be expressed as a linear 
combination of $\{ \v_c^t \}_{c=1}^C$, i.e.,
\begin{equation}
\label{eq:ps0a}
\a_i^t = \V^t \u_i^t + \e_i^t, \;\; i = 1, \dots, N
\end{equation}
where $ \left( \a_i^t \right)^{\top} $ denotes row $i$ of $\A^t$, and 
$\e_i^t$ captures unmodeled dynamics. Entries of 
$\u_i^t := [u_{i1}^t \dots u_{iC}^t]^{\top} \in \mathbb{R}^{C}$ 
in \eqref{eq:ps0a} assign community affiliation strengths, with 
$u_{ic}^t = 0$ only if node $i$ does not belong to community
$c$ during interval $t$. Since $a_{ij}^t \geq 0$, entries of 
both $\V^t$ and $\u_i^t$ will be constrained to nonnegative
values. Collecting edge vectors for all nodes, \eqref{eq:ps0a} 
can be expressed equivalently as the following canonical
NMF model
\begin{equation}
\label{eq:ps0b}
\A^t = \U^t(\V^t)^{\top} + \E^t
\end{equation}
where $(\U^t)^{\top} := [\u_1^t \dots \u_N^t]$, and
$(\E^t)^{\top} := [\e_1^t \dots \e_N^t]$.
The asymmetry inherent to \eqref{eq:ps0b}
generalizes traditional approaches (e.g., 
spectral clustering), rendering them capable of capturing 
communities in weighted, directed, and even bipartite graphs. The
special case in which edges are undirected (i.e., $a_{ij}^t = a_{ji}^t$)
can be readily realized $\U^t \equiv \V^t$. 

Contemporary community detection approaches
overwhelmingly focus on unipartite networks, whereby edges are allowed
to exist between any pair of nodes. On the other hand, directional edges in 
bipartite networks only connect nodes belonging to two distinct
classes. Examples include buyer-seller networks existing in online 
applications like E-bay,
or recommender networks with edges capturing ratings of products by 
customers. With matrices $\U^t \in \mathbb{R}^{N \times C}$, and 
$\V^t \in \mathbb{R}^{M \times C}$, communities in an $NM$-node bipartite
network ($N$ nodes belonging to one class, and $M$ to the other)
can be readily captured through the affiliation model~\eqref{eq:ps0b}.

Unfortunately, \eqref{eq:ps0a} does not effectively
capture anomalous nodes that exhibit unusually strong
affiliation to one or more communities. 
This aberrant behavior has been observed
in several real-world networks, and it often arises due  
to any of several reasons. For example, the presence of
``fake'' user accounts in online social networks created for
phishing purposes from unsuspecting peers may lead to abnormally
high numbers of outgoing links. In addition, fraudulent 
reviewers in web-based rating applications may exhibit abnormal
affiliation while unfairly promoting their services to specific
communities. Regardless of the underlying reason, detection of such
nodes is envisioned as a source of strategic information for network 
operators. Moreover, identification of anomalies is critical for 
improved community detection accuracy; see Figure~\ref{fig:anets} for 
an example where an anomalous node distorts the underlying
community structure.

\subsection{Outlier-aware community affiliation model}
\label{ssec:ocamodel}
Suppose node $i$ is considered anomalous, exhibiting an 
abnormally strong level of affiliation in one or more of 
the $C$ communities. In order to preserve the estimation accuracy of
community discovery algorithms,
one is motivated to modify \eqref{eq:ps0a} so that such outliers are
accounted for. The present paper postulates the following robust 
edge generation model
\begin{equation}
\label{eq:ps11}
\a_i^t = \V^t (\u_i^t + \o_i^t) + \e_i^t, \;\; i = 1, \dots, N
\end{equation}
where $\o_i^t := [o_{i1}^t \dots o_{iC}^t]^{\top}$, and $o_{ic}^t \neq 0$ 
only if node $i$ exhibits anomalous
affiliation in community $c$. Intuitively, \eqref{eq:ps11} reasonably 
suggests that one can investigate whether any node is an anomaly or 
not by introducing more variables that compensate for the effect of outliers
on the edge generation model. Since outliers are rare by definition,
vectors $\{ \o_i^t \}_{i=1}^N$ are generally sparse, and this prior
knowledge can be exploited to recover the unknowns.

Letting $(\O^t)^{\top} := [\o_1^t \dots \o_N^t]$, 
the outlier-aware community affiliation 
model in~\eqref{eq:ps0b} can be written as 
\begin{equation}
\label{eq:ps3}
\A^t = (\U^t + \O^t) (\V^t)^{\top} + \E^t, \;\; t = 1, 2, \dots.
\end{equation}
where $\O^t$ is sparse. 
In static scenarios with $\A = (\U + \O) \V^{\top}$, setting 
$\O = \mathbf{0}$ yields the canonical NMF model for community discovery. 
Given $\{ \A^t \}_{t=1}^T$, the goal of this paper 
is to track the community affiliation matrices 
$\{ \U^t, \V^t \}_{t=1}^T$, as well as outliers 
captured through matrices $\{ \O^t \}_{t=1}^T$.

It is worth reiterating that \eqref{eq:ps3} is a
heavily under-determined model, and the only hope to
recover $\{ \U^t, \O^t, \V^t \}_{t=1}^T$ lies
in exploiting prior information about the 
structure of the unknowns. Indeed, the estimator
advocated in the sequel will 
capitalize on sparsity, low rank, and the slow evolution
of networks. Introducing extra variables to capture outliers has been used 
in different contexts; see e.g.,~\cite{gg} and references therein.  

\begin{remark}[Measurement outliers]\label{rem:meas}
\normalfont Model \ref{eq:ps3}
is motivated by anomalous nodes, whose presence leads to distorted
community structures in e.g., social networks. A slight variation of 
this problem arises in cases where one is interested in identifying
which edges are anomalous. This is well-motivated in settings where edge weights
are directly measured, and encode valuable information e.g., star ratings in 
online review systems. The outliers in such cases are the result of 
faulty measurements, bad data (e.g., skewed user ratings), or data corruption.
To detect anomalous edge weights, one can postulate that [cf. \eqref{eq:ps3}]
\begin{equation}
\label{eq:rps1}
\A^t = \U^t  (\V^t)^{\top} + \O^t + \E^t, \;\; t = 1, 2, \dots.
\end{equation}
where $\O^t $ is sparse, and can be effectively recovered
by leveraging sparsity-promoting estimation approaches; 
see e.g.,~\cite{gg} and references therein.
\end{remark}

\section{Community Tracking Algorithm}
\label{sec:ctalg}
This section assumes that the following hold:
a1) $\O^t $ is sparse; a2) $\U^t(\V^t)^{\top}$ is low rank; and 
a3) $\{ \A^t \}_{t=1}^T$ evolve slowly over time, that is, the sequence of 
matrices $ \{ \A^t - \A^{t-1} \}_{t=1}^T $ are sparse. 
In order to justify a1, note that the set of anomalous nodes
is much smaller than that of ``ordinary'' nodes.
On the other hand, a2 results from requiring that $\text{rank}(\U^t(\V^t)^{\top}) \leq C \ll N$,
while a3 is motivated by observations of the evolution of most real-world networks.
In the sequel, a sequential estimator that exploits a1-a3 is put forth.

\subsection{Exponentially-weighted least-squares estimator}
Suppose data are acquired sequentially over time, and 
storage memory is limited; thus, it is impractical to
aim for batch estimation. Under the
aforementioned assumptions, the following sparsity-promoting,
rank-regularized, and exponentially-weighted  least-squares (EWLS) 
estimator can track the unknown matrices
\begin{multline}
\label{eq:cta1}
\{\hat{\U}^t, \hat{\V}^t, \hat{\O}^t  \} \\ 
= \underset{ \left\lbrace \U, \V, \O \right\rbrace \in \mathbb{R}_{+}^{N \times C} }{\text{arg min}}  \;\;
\sum\limits_{\tau = 1}^t \beta^{t-\tau} \big \| \A^{\tau} 
- (\U + \O) \V^{\top} \big \|_{F}^2 
\\ + \lambda_t \big \| \U \V^{\top} \big \|_{\ast}
+ \mu_t \big \| \O  \big \|_0
\end{multline}
where the nuclear norm $\| \U \V^{\top} \|_{\ast} := \sum_{n} \sigma_n (\U \V^{\top})$
sums the singular values of $\U \V^{\top}$, while
$ \| \O \|_0 := \sum_{ik} \mathbb{1}_{ \{ o_{ik} \neq 0 \} } $ counts
the non-zero entries in $\O$. Regularization parameters
$\lambda_t \geq 0$ and $\mu_t \geq 0$ control the low rank
of $\U \V^{\top}$ and sparsity in $\O$, respectively. Finally,
$\beta^{t-\tau}$  is a ``forgetting'' factor with $\beta \in ( 0, 1 ] $,
which facilitates tracking slow variations by down-weighing
past data when $\beta < 1$.

Problem \eqref{eq:cta1} is non-convex and NP-hard
to solve. Nevertheless, this can be circumvented by resorting to 
tight convex relaxation. Specifically, $\| \O \|_0$ can be surrogated with
$\| \O \|_1 := \sum_{ic} |o_{ic}|$, and one can leverage the following
characterization of the nuclear norm; see e.g.,~\cite{morteza_dist} 
\begin{equation}
\label{eq:cta2}
\big \| \Z  \big \|_{\ast} := \underset{\U, \V \in \mathbb{R}^{N \times C}}{\text{min}}
\; \frac{1}{2} \bigg \{ \big \| \U  \big \|_{F}^2 +  \big \| \V  \big \|_{F}^2 \bigg \} \;\; \text{s.t.} \;\; \Z = \U \V^{\top}
\end{equation}
which leads to the following optimization problem
\begin{multline}
\label{eq:cta3}
\{\hat{\U}^t, \hat{\V}^t, \hat{\O}^t  \}  \\ 
= \underset{ \left\lbrace \U, \V, \O \right\rbrace \in \mathbb{R}_{+}^{N \times C} }{\text{arg min}}  \;\;
\sum\limits_{\tau = 1}^t \beta^{t-\tau} \big \| \A^{\tau} 
- (\U + \O) \V^{\top} \big \|_{F}^2 \\
  + \frac{\lambda_t}{2} \bigg \{ \big \| \U \big \|_F^2 + \big \| \V  \big \|_F^2 \bigg \}  
+ \mu_t \big \| \O \big \|_1
\end{multline}
whose separability renders it amenable to alternating minimization (AM) strategies, 
as discussed next.

\subsection{Alternating minimization}
\label{subsec:am}
Focusing on the first term of the cost in~\eqref{eq:cta3}, note that
\begin{multline}
\label{eq:am1}
\sum\limits_{\tau = 1}^t \beta^{t-\tau} \big \| \A^{\tau} - (\U + \O) \V^{\top} \big \|_{F}^2
 \\
= s^t \text{Tr} \bigg\{ \V(\U^{\top}\U + \O^{\top}\O)\V^{\top} + 2\V \U^{\top}\O\V^{\top} \bigg\} \\
- 2 \text{Tr} \bigg\{ (\S^{t})^{\top}(\U + \O)\V^{\top}  \bigg\}
\end{multline}
where $s^t := (1 - \beta^{t})/(1 - \beta)$ and $\S^t := \A^t + \beta \S^{t-1}$
recursively accumulate past data with minimal storage
requirements. Since~\eqref{eq:cta3} is separable across the optimization
variables, one can resort to iterative AM, by 
alternately solving for each variable while holding the others fixed. With 
\begin{equation}
\label{eq:am_1a}
\Phi \left( \U, \V, \O, \S^t, s^t \right) := 
\sum\limits_{\tau = 1}^t \beta^{t-\tau} \big \| \A^{\tau} - (\U + \O) \V^{\top} \big \|_{F}^2
\end{equation}
AM iterations amount to the following per-interval updates
\begin{subequations}
\label{subeq:am}
\begin{eqnarray}
\label{eq:am2}
\nonumber
\U[k]  &=&  \underset{\U \in \mathbb{R}_{+}^{N \times C}}{\text{arg min}} \;\;\;
\Phi \left( \U, \V[k-1], \O[k-1], \S^t, s^t \right) \\
& &  + \left( \lambda_t/2 \right) \big \| \U \big \|_F^2 \\
\label{eq:am3}
\nonumber
\V[k] &=& \underset{\V \in \mathbb{R}_{+}^{N \times C}}{\text{arg min}} \;\;\;
\Phi \left( \U[k], \V, \O[k-1], \S^t, s^t \right) \\
& & + \left( \lambda_t/2 \right) \big \| \V \big \|_F^2 \\
\label{eq:am4}
\nonumber
\O[k] &=& \underset{\O \in \mathbb{R}_{+}^{N \times C}}{\text{arg min}} \;\;\;
\Phi \left( \U[k], \V[k], \O, \S^t, s^t \right) \\
& &  + \mu_t \big \| \O \big \|_1
\end{eqnarray}
\end{subequations}
over iterations indexed by $k$, until convergence is achieved.
%
%
\begin{algorithm}
    \caption{Alternating minimization}
\label{alg1}
\begin{algorithmic}[1]
   \STATE {\bfseries Input:}  $\{ \A^{t} \}_{t=1}^{T}$, $\beta$,  $C$
   \STATE Initialize $\U[0]$, $\V[0]$, $\O[0]$
   \STATE Set $\S^0 = \mathbf{0}$
   \FOR{$t=1 \dots $ }
   \STATE Set $s^t = (1-\beta^t)/(1-\beta)$
   \STATE Update $\S^t = \A^t + \beta \S^{t-1}$
   \STATE Initialize $k = 0$
   \REPEAT
   \STATE $k = k + 1$
   \STATE Set $ \{\alpha_{u,k}, \alpha_{v,k} \}$
   \STATE Compute $\U[k]$ via \eqref{eq:am5}
   \STATE Compute $\V[k]$ via \eqref{eq:am6}
   \STATE $r = 0, \W_{r}[k] = \O_r[k] = \O[k-1], \theta_r[k] = 1$
   \REPEAT
   \STATE Update $\X_r[k]$ via \eqref{eq:fst2b}
   \STATE $ \O_{r} [k]  = \left[ \mathcal{S}_{\mu_t/L_{\Psi}} \left(  \X_{r}[k]  \right) \right]_{+}$
   \STATE $\theta_{r+1}[k]~=~(1~+~\sqrt{1 + 4\theta_{r}^2[k]})/2$
   \STATE Update $\W_{r+1}[k]$ via \eqref{eq:fst3}
   \STATE $r = r + 1$
   \UNTIL $\O_{r}[k] $ converges
   \STATE $\O[k] = \O_r[k]$
   \UNTIL $\{ \U[k], \V[k], \O[k] \}$ converge
   \STATE $\hat{\U}^t = \U[k], \hat{\V}^t = \V[k], \hat{\O}^t = \O[k]$
   \ENDFOR
\end{algorithmic}
\end{algorithm}
%

The constrained subproblems in \eqref{eq:am2} and \eqref{eq:am3} are convex, and
can be readily solved via projected gradient (PG) iterations. 
Since the gradients of their cost functions are available, and the projection 
operator onto the nonnegative orthant is well defined, PG iterations 
are guaranteed to eventually converge to the optimal solution~\cite[p.~223]{bert}. 
Per iteration $k$, PG updates amount to setting
\begin{multline}
\label{eq:am5}
\U [k] = \big[ \U[k-1]  - \alpha_{u,k} \nabla_{\U} \Phi \big(\U[k-1],\\ 
 \V[k-1], \O[k-1], \S^t, s^t \big) \big]_{+}
\end{multline}
\begin{multline}
\label{eq:am6}
\V[k] = \big[ \V[k-1]  - \alpha_{v,k} \nabla_{\V} \Phi \big( \U[k], \\
\V[k-1], \O[k-1], \S^t, s^t \big) \big]_{+}
\end{multline}
where $\alpha_{u,k}$ and $\alpha_{v,k}$ denote (possibly) iteration-dependent
step sizes. 
In addition, $\nabla_{\U} \Phi(.) \left( \nabla_{\V} \Phi(.) \right)$ 
denotes the gradient of $\Phi(.) $ 
with respect to $\U \left( \V \right)$. Expressions for the gradients
can be readily obtained, but are omitted here due to space constraints. 

The cost function in~\eqref{eq:am4} is convex with both 
smooth and non-smooth terms. Recent advances in proximal algorithms
have led to efficient, provably-convergent iterative schemes 
for solving such optimization problems.
We will resort to the \emph{fast iterative shrinkage thresholding algorithm (FISTA)}
whose accelerated convergence rate renders it attractive for 
sequential learning~\cite{beck}.

\subsection{FISTA for outlier updates}
\label{ssec:fista}
Note that \eqref{eq:am4} does not admit a closed-form solution,
and the proposed strategy will entail a number of 
inner iterations (indexed by $r$), per AM iteration $k$.
FISTA solves for $\O$ in \eqref{eq:am4} through a two-step update
involving gradient descent on $\Phi(.)$,
evaluated at a linear combination of the two most recent iterates,
followed by a closed-form soft-thresholding step per
iteration $r$. Setting $\theta_{0}[k] = 1$ and $\W_{1}[k] = \O_{k-1}$, 
it turns out that the updates can be written as~\cite{beck}
\begin{equation}
\label{eq:fst2}
\O_{r}[k] = \underset{\O \in \mathbb{R}_{+}^{N \times C}}{\text{arg min}}
\left(L_{\Phi}/2\right) \big \| \O - \X_{r}[k] \big \|_F^2 + \mu_t \big \| \O  \big \|_1
\end{equation}
where 
\begin{multline}
\label{eq:fst2b}
\X_{r}[k] = ( \W_{r}[k] \\ - (1/L_{\Phi}) 
\nabla_{\O} \Phi(\W_{r}[k], \U[k], \V[k], \S^t, s^t))
\end{multline}
with
\[ \theta_{r+1}[k]~=~(1~+~\sqrt{1 + 4\theta_{r}^2[k]})/2. \]
Furthermore,
\begin{equation}
\label{eq:fst3}
\W_{r+1}[k] = \O_{r}[k] + \left( \frac{\theta_{r}[k] - 1}{\theta_{r+1}[k]} \right) (\O_{r}[k] - \O_{r-1}[k] )
\end{equation}
where $L_{\Phi}$ denotes a \emph{Lipschitz} constant of
$\nabla_{\O} \Phi(.)$.
Note that~\eqref{eq:fst2} is similar to the so-termed \emph{
least-absolute shrinkage and selection operator (Lasso)}
with a closed-form solution, namely
$\left[ \O_{r}[k] \right]_{ij} = \left[ \mathcal{S}_{\mu_t/L_{\Phi}} 
 \left( [ \X_{r}[k] ]_{ij} \right) \right]_{+}$,
where the thresholding operator is defined entry-wise
as $\mathcal{S}_{ \mu }(x) := ( |x| - \mu )_{+} \text{sign}(x) $~\cite[Ch.~3]{hastie_book}.
Computation of $\O[k]$ entails solving~\eqref{eq:fst2} over several iterations
indexed by $r$ until convergence. Algorithm~\ref{alg1} summarizes
the details of the developed community tracking scheme, with 
$\beta$ and $C$ assumed to be given as algorithm inputs. 

\section{Delay-sensitive operation}
\label{sec:delay}
Algorithm~\ref{alg1} relies upon convergence of the unknown
variables per time interval. Unfortunately, this mode of operation
is not suitable for delay-sensitive applications, where decisions must
be made ``on the fly.'' In fact, one may be willing
to trade off solution accuracy for real-time operation in certain
application domains. This section puts forth a couple of algorithmic
enhancements that will facilitate real-time tracking, namely
by premature termination of PG iterations, and leveraging 
the stochastic gradient descent framework.

\subsection{Premature termination}
\label{ssec:premat}
For networks that generally
evolve slowly over time, it is not necessary
to run the tracking algorithm until convergence per
time interval. Since a compromise can be struck between
an accurate solution computed slowly
and a less accurate solution computed very fast, one can judiciously truncate
the number of inner iterations to $k_{\text{max}}$.
This premature termination is well motivated when
the network topology is piecewise stationary with
sufficiently long coherence time, with respect to the 
number of time intervals. It is then unnecessary to
seek convergence per time interval since it can be argued
that a ``good'' solution will be attained across time intervals
before the topology changes. If the network topology varies 
in accordance with a stationary distribution, it can 
be demonstrated that convergence will eventually be attained, even when
$k_{\text{max}} = 1$ i.e., running a single iteration 
per time interval. Algorithm~\ref{alg2} summarizes the steps
involved in this inexact tracking scheme under the special case
with $k_{\text{max}} = 1$.
%
%
\begin{algorithm}
    \caption{Inexact alternating minimization}
\label{alg2}
\begin{algorithmic}[1]
   \STATE {\bfseries Input:}  $\{ \A^{t} \}_{t=1}^{T}$, $\beta$,  $C$
   \STATE Initialize $\U^0$, $\V^0$, $\O^0$
   \STATE Set $\S^0 = \mathbf{0}$, $\W^{0} = \O^0, \theta_0 = 1$
   \FOR{$t=1 \dots T$ }
   \STATE Set $s^t = (1-\beta^t)/(1-\beta)$
   \STATE Update $\S^t = \A^t + \beta \S^{t-1}$

   \STATE Set $ \{\alpha_{u,t}, \alpha_{v,t} \}$
   \STATE $\U^t = \big[ \U^{t-1} - \alpha_{u,t} \nabla_{\U} \Phi(\U^{t-1}, \V^{t-1}, \O^{t-1}, \S^t, s^t) \big]_{+}$
   \STATE $\V^{t} = \big[ \V^{t-1} - \alpha_{v,t} \nabla_{\V} \Phi(\V^{t-1}, \U^{t}, \O^{t-1}, \S^t, s^t) \big]_{+}$
   \STATE $\X^{t} = ( \W^{t} - \frac{1}{L_{\Phi}} \nabla_{\O} \Phi(\W^{t}, \U^{t}, \V^t, \S^t, s^t))$
   \STATE $ \O^{t}  = \left[ \mathcal{S}_{\mu_t/L_{\Phi}} \left(  \X^{t}  \right) \right]_{+}$
   \STATE $\theta_{t}~=~(1~+~\sqrt{1 + 4\theta_{t-1}^2})/2$
   \STATE $\W^{t} = \O^{t} + \left( (\theta_{t-1} - 1)/ \theta_{t} \right) (\O^{t} - \O^{t-1} )$
   \ENDFOR
\end{algorithmic}
\end{algorithm}

\subsection{Stochastic gradient descent}
\label{ssec:sgd}
Instead of premature termination, real-time operation 
can be facilitated by resorting to stochastic gradient
descent (SGD) iterations. Let 
\begin{equation}
\label{eq:sgd0a}
\zeta_{\tau} \left( \U, \V, \O, \A^{\tau} \right) := 
\big \| \A^{\tau} - (\U + \O) \V^{\top}  \big \|_F^2
\end{equation}
and 
\begin{equation}
\label{eq:sgd0b}
\zeta_{\lambda_{\tau}, \mu_{\tau}}  \left( \U, \V, \O \right) := 
\frac{\lambda_{\tau}}{2} \left\lbrace \big \| \U  \big \|_F^2 + \big \| \V  \big \|_F^2 \right \rbrace
+ \mu_{\tau} \big \| \O \big \|_1
\end{equation}
and consider the online learning setup, where the goal is to minimize
the expected cost $E \{ \zeta_{\tau} \left( \U, \V, \O, \A^{\tau} \right) + 
\zeta_{\lambda_{\tau}, \mu_{\tau}}  \left( \U, \V, \O \right) \}$ (with respect to
the unknown probability distribution of the data). The present paper
pursues an online learning strategy, in which the expected cost is replaced 
with the empirical cost function
$ (1/t)  \sum_{\tau = 1}^t \left[ \zeta_{\tau} \left( \U, \V, \O, \A^{\tau} \right) + 
\zeta_{\lambda_{\tau}, \mu_{\tau}}  \left( \U, \V, \O \right) \right]$ as a surrogate.
Generally, SGD is applicable 
to separable sum-minimization settings, in which the $\tau$-th 
summand is a function of the $\tau$-th datum.
To incur the least computational and memory storage costs, the SGD
approach advocated here discards all past data, and solves 
\begin{equation}
\label{eq:sgd0c}
\underset{\U, \V, \O \in \mathbb{R}_{+}^{N \times C}}{\text{arg min}} \;\;
\zeta_t \left( \U, \V, \O, \A^t \right) + 
\zeta_{\lambda_t, \mu_t}  \left( \U, \V, \O \right)
\end{equation}
per interval $t$, which is tantamount to solving the EWLSE upon setting
$\beta = 0$. This tracking scheme is reminiscent of the popular \emph{least mean-squares (LMS)}
algorithm, that has been well-studied within the context of
adaptive estimation; see e.g.,~\cite{solo_book}.  
With $\mathbf{Y} := [\U \; \V \; \O]$, a common approach
to solving~\eqref{eq:sgd0c} involves the projected iteration (see e.g., \cite{chen})
\begin{equation}
\label{eq:sgd0d}
\mathbf{Y}^t = \left[ \mathbf{Y}^{t-1} - \eta \partial \left( \zeta_t (\mathbf{Y}) + \zeta_{\lambda_t, \mu_t} (\mathbf{Y}) \right) \big |_{\mathbf{Y} = \mathbf{Y}^{t-1}} \right]_+
\end{equation}
where $\partial(.)$ denotes the subgradient operator, and $\eta \ge 0$ is a small
constant step-size. A major limitation associated with \eqref{eq:sgd0d} 
is that the resulting per-iteration solutions $\O^t$ are generally not 
sparse. Instead of subgradients, an AM procedure with FISTA updates
adopted for $\O^t$ is followed in a manner similar to Algorithms~\ref{alg1}
and~\ref{alg2}. The main differences stem from eliminating the recursive updates,
and adopting constant step sizes to facilitate tracking.

To this end, the following subproblems are solved during interval $t$,
\begin{multline}
\label{eq:sgd2}
\U^t = \underset{\U \in \mathbb{R}^{N \times C}_{+}}{\text{arg min}} \;
\zeta_t \left( \U, \V^{t-1}, \O^{t-1}, \A^t \right) \\
 + \zeta_{\lambda_t, \mu_t}  \left( \U, \V^{t-1}, \O^{t-1} \right)
\end{multline}
\begin{multline}
\label{eq:sgd3}
\V^t = \underset{\V \in \mathbb{R}^{N \times C}_{+}}{\text{arg min}} \;
\zeta_t \left( \U^t, \V, \O^{t-1}, \A^t \right)\\
 + \zeta_{\lambda_t, \mu_t}  \left( \U^t, \V, \O^{t-1} \right)
\end{multline}
\begin{multline}
\label{eq:sgd4}
\O^t = \underset{\O \in \mathbb{R}^{N \times C}_{+}}{\text{arg min}} \;
\zeta_t \left( \U^t, \V^t, \O, \A^t \right) \\ 
+ \zeta_{\lambda_t, \mu_t}  \left( \U^t, \V^t, \O \right).
\end{multline}
In order to operate in real-time, $\U$ and $\V$ can be updated by a single
gradient descent step per time slot. Similarly, minimization of
\eqref{eq:sgd4} across time slots entails a single FISTA update 
that linearly combines $\O^{t-1}$ and $\O^{t-2}$. Exact
algorithmic details of the SGD-based community tracking algorithm
are tabulated in Algorithm~\ref{alg3}, with $\Upsilon_t (\U, \V, \O, \A^t)
:= \zeta_t \left( \U, \V, \O, \A^t \right) + 
\zeta_{\lambda_t, \mu_t}  \left( \U, \V, \O \right)$.
%
%
\begin{algorithm}
    \caption{SGD tracking algorithm}
\label{alg3}
\begin{algorithmic}[1]
   \STATE {\bfseries Input:}  $\{ \A^{t} \}_{t=1}^{T}$, $\beta$,  $C, \alpha_u, \alpha_v$
   \STATE Initialize $\U^0$, $\V^0$, $\O^0$
   \STATE Set $\W^{0} = \O^0, \theta_0 = 1$
   \FOR{$t=1 \dots T$ }
   \STATE $\U^t = \big[ \U^{t-1} - \alpha_u \nabla_{\U}\Upsilon_t (\U, \V^{t-1}, \O^{t-1}, \A^t) \big]_{+}$
   \STATE $\V^{t} = \big[ \V^{t-1} - \alpha_v \nabla_{\V} \Upsilon_t (\U^t, \V, \O^{t-1}, \A^t) \big]_{+}$
   \STATE $\X^{t} = ( \W^{t} - \frac{1}{L_{\Phi}} \nabla_{\O} \Phi(\W^{t}, \U^{t}, \V^t, \A^t, 1))$
   \STATE $ \O^{t}  = \left[ \mathcal{S}_{\mu_t/L_{\Phi}} \left(  \X^{t}  \right) \right]_{+}$
   \STATE $\theta_{t}~=~(1~+~\sqrt{1 + 4\theta_{t-1}^2})/2$
   \STATE $\W^{t} = \O^{t} + \left( (\theta_{t-1} - 1)/ \theta_{t} \right) (\O^{t} - \O^{t-1} )$
   \ENDFOR
\end{algorithmic}
\end{algorithm}

\section{Decentralized Implementation}
\label{sec:decent}
The tracking algorithms developed so far have assumed that
connectivity data (i.e., $\{ \A^t \}$) are acquired and processed in a centralized
fashion. This may turn out to be infeasible, since for example, 
certain applications store large graphs over distributed file storage 
system hosted across a large network of computers. 
In fact, graphs capturing the web link structure, 
and online social networks are typically stored as  
``chunks'' of files that are both distributed across computing
nodes, and spatially over several geographical sites. 
In addition to the inherent computational bottlenecks, soaring
data communication costs would render centralized approaches 
infeasible in such scenarios.

In lieu of these computational constraints, this section 
puts forth a decentralized algorithm that jointly tracks
temporal communities and anomalous nodes. 
The alternating-direction method of multipliers (ADMM)
has recently emerged as powerful tool for decentralized optimization
problems~\cite{schizas3}, and it will be adopted here for the 
community tracking task. A connected network of computing agents 
is deployed, with links representing direct communication paths 
between nodes. The key idea is that each node iteratively 
solves the problem using only a subset of the input data, while
exchanging intermediate solutions with single-hop neighbors 
until consensus is achieved.

\subsection{Consensus constraints}
\label{ssec:cons}
Consider an undirected graph $\mathcal{G} = (\mathcal{M}, \mathcal{L})$ whose
vertices $\mathcal{M} := \{ 1,\dots,M \}$ are $M$ spatially-distributed computing
agents, and whose edges $\mathcal{L} := \{1,\dots,L\}$ are representative
of direct communication links between pairs of agents. It is assumed that the 
processing network abstracted by $\mathcal{G}$ is connected, so that
(multi-hop) communication is possible between any pair of agents. Suppose the temporal
adjacency matrix is partitioned as follows $\A^t := [(\A_1^t)^{\top}, \dots, (\A_M^t)^{\top}]^{\top}$,
and it is distributed across the processing network. During
time interval $t$, agent $m$ receives the submatrix $\A_m^t$.
To minimize communication costs, each agent is only allowed
to send and receive data from its single-hop neighborhood $\mathcal{N}_m$.
Let $\U := [\U_1^{\top}, \dots, \U_M^{\top}]^{\top}$ and 
$\O := [\O_1^{\top}, \dots, \O_M^{\top}]^{\top}$, where 
$\U_m \in \mathbb{R}^{N_m \times C}$, $\O_m \in \mathbb{R}^{N_m \times C}$,
and $\sum_{m=1}^M N_m = N$. In terms of the per-agent submatrices,
\eqref{eq:cta3} can now be written as follows
\begin{multline}
\label{eq:admm1}
\underset{ \left\lbrace \substack{ \{ \U_m, \O_m \in \mathbb{R}^{N_m \times C}_{+} \}_{m=1}^{M} \\
\V \in \mathbb{R}^{N \times C}_{+}} \right\rbrace }{\text{arg min}} 
\sum\limits_{m=1}^{M} \bigg[   \sum\limits_{\tau = 1}^t \beta^{t-\tau} \big \| \A_m^{\tau}  
- (\U_m  + \O_m)\\
\times  \V^{\top} \big \|_{F}^2  + \frac{\lambda_t}{2} \big \| \U_m \big \|_F^2 
+ \frac{\lambda_t}{2M} \big \| \V  \big \|_F^2   + \mu_t \big \| \O_m  \big \|_1 \bigg]
\end{multline}
for $t = 1,\dots,T$. Clearly $\U$ and $\O$ decouple
across computing agents, whereas $\V$ does not. A viable approach entails
allowing each agent to solve for its corresponding unknowns $\U_m$
and $\O_m$ in parallel, followed by communication
of the estimates to a central processing node that solves for 
$\V$. The downside to this approach is that it involves a significant 
communication and storage overhead as the central node must
receive intermediate values of $\{ \U_m, \O_m \}_{m=1}^M$, and 
then broadcast its computed value of $\V$ per iteration. In addition,
this introduces the risk of a single point of failure at the central
node.

To operate in a truly decentralized manner, each agent
will solve for $\V$ independently under \emph{consensus} 
constraints requiring equality of the solution to those 
computed by single-hop neighbors~\cite{schizas3}. Since $\mathcal{G}$
is connected, it can be readily shown that consensus
on the solution for $\V$ will be reached upon convergence of
the algorithm~\cite{schizas3}. Incorporating consensus constraints,
each time interval entails solving the following fully
decoupled minimization problem
\begin{eqnarray}
\label{eq:admm2}
\nonumber
\underset{ \left\lbrace  \U_m, \O_m, \V_m \right\rbrace_{m=1}^{M} }{\text{arg min}} &
\sum\limits_{m=1}^{M} \bigg[ \Phi \left( \U_m, \V_m, \O_m, \S_m^t, s^t \right)  \\
\nonumber
&  + \frac{\lambda_t}{2} \big \| \U_m \big \|_F^2 
  + \frac{\lambda_t}{2M} \big \| \V_m \big \|_F^2   + \mu_t \big \| \O_m \big \|_1 \bigg] \\
 \nonumber
 \text{s. to} & \{ \U_m, \O_m \} \in \mathbb{R}_{+}^{N_m \times C}, \V_m \in \mathbb{R}_{+}^{N \times C} \\
 &   \V_m = \V_n, n \in \mathcal{N}_m.
\end{eqnarray}
Letting 
\begin{multline}
\label{eq:admm2a}
\Psi_{\lambda_t}(\U_m, \O_m, \V_m, \S_m^t, s^t) := 
\Phi \left( \U_m, \V_m, \O_m, \S_m^t, s^t \right) \\ 
+ \frac{\lambda_t}{2} \| \U_m \|_F^2
 + \frac{\lambda_t}{2M} \| \V_m \|_F^2
\end{multline}
and introducing dummy variables $\{ \P_m \}_{m=1}^M$,
\eqref{eq:admm2} can be written as follows
\begin{eqnarray}
\label{eq:admm3}
\nonumber
\underset{ \left\lbrace  \U_m, \O_m, \V_m \right\rbrace_{m=1}^{M} }{\text{arg min}} & 
\sum\limits_{m=1}^{M} \bigg[ \Psi_{\lambda_t}(\U_m, \O_m, \V_m, \S_m^t, s^t) \\ 
\nonumber 
& + \mu_t \| \P_m \|_1 \bigg] \\
 \nonumber
  \text{s. to} & \{ \U_m, \O_m \} \in \mathbb{R}_{+}^{N_m \times C}, \V_m \in \mathbb{R}_{+}^{N \times C} \\
 &  \O_m = \P_m, \V_m = \V_n, n \in \mathcal{N}_m.
\end{eqnarray}
In order to solve \eqref{eq:admm3}, introduce the variables $\{ \bar{\X}_{nm}, \tilde{\X}_{nm} \}$,
and modify the constraints $\V_m = \V_n, n \in \mathcal{N}_m$ as
\[
\V_m = \bar{\X}_{nm}, \V_n = \tilde{\X}_{nm}, \bar{\X}_{nm} = \tilde{\X}_{nm}, n \in \mathcal{N}_m.
\]
Introducing the dual variables $\{\boldsymbol{\Gamma} _m\}_{m=1}^M$, and $\{ \{ \bar{\boldsymbol{\Pi}}_{nm}, \tilde{\boldsymbol{\Pi}}_{nm} \}_{n \in \mathcal{N}_m} \}_{m=1}^M$, and temporarily ignoring 
non-negativity constraints, the resulting augmented Lagrangian can be written as
\begin{multline}
\label{eq:admm4}
\mathcal{L}_{\rho}(\mathcal{P}_1, \mathcal{P}_2, \mathcal{P}_3, \mathcal{D}  ) = 
\sum\limits_{m=1}^{M} \bigg[ \Psi_{\lambda_t}(\U_m, \O_m, \V_m, \S_m^t, s^t) \\ 
+ \mu_t \| \P_m \|_1 \bigg]
    + \sum\limits_{m=1}^{M} \bigg[ \text{Tr}\left(\boldsymbol{\Gamma}_m^{\top} (\O_m - \P_m)\right) \\
   + \sum\limits_{n \in \mathcal{N}_m} \text{Tr}\left\lbrace \bar{\boldsymbol{\Pi}}_{nm}^{\top} (\V_m - \bar{\X}_{nm})  +  \tilde{\boldsymbol{\Pi}}_{nm}^{\top} (\V_n - \tilde{\X}_{nm})   \right\rbrace \bigg] \\
 + \frac{\rho}{2} \sum\limits_{m=1}^{M} \bigg[ \| \O_m - \P_m \|_F^2 \\
   + \sum\limits_{n \in \mathcal{N}_m} \left\lbrace \| \V_m - \bar{\X}_{nm} \|_F^2  + 
   \| \V_n - \tilde{\X}_{nm} \|_F^2 \right\rbrace \bigg]
\end{multline}
where $\rho > 0$, $\mathcal{P}_1 := \{ \V_m \}_{m=1}^M$, $\mathcal{P}_2 := \{ \U_m, \O_m, \P_m \}_{m=1}^M$, and $\mathcal{P}_3 := \{ \{ \bar{\X}_{nm}, \tilde{\X}_{nm} \}_{n \in \mathcal{N}_m} \}_{m=1}^M$ denote primal variables, while 
$\mathcal{D} := \{ \boldsymbol{\Gamma}_m, \{ \bar{\boldsymbol{\Pi}}_{nm}, \tilde{\boldsymbol{\Pi}}_{nm} \}_{n \in \mathcal{N}_m} \}_{m=1}^M$
denotes the set of dual variables.

Towards applying ADMM to \eqref{eq:admm4}, an iterative strategy is pursued, 
entailing dual variable update as the first step, followed
by alternate minimization of $\mathcal{L}_{\rho}(\mathcal{P}_1, \mathcal{P}_2, \mathcal{P}_3, \mathcal{D}  )$ over each of the primal variables, while holding the rest fixed to their most recent values. Since $\mathcal{L}_{\rho}(\mathcal{P}_1, \mathcal{P}_2, \mathcal{P}_3, \mathcal{D}  )$ is
completely decoupled across the $M$ computing agents, the problem can be solved in 
an entirely decentralized manner. The per-agent updates of the proposed algorithm during iteration $k$ comprise the following steps.

\textbf{[S1]: Dual variable update.}
\begin{subequations}
\begin{eqnarray}
\label{eq:admm5a}
\boldsymbol{\Gamma}_m[k+1] &=& 
\boldsymbol{\Gamma}_m[k] + \rho (\O_m[k] - \P_m[k]) \\
\label{eq:admm5b}
\bar{\boldsymbol{\Pi}}_{nm}[k+1] &=& 
\bar{\boldsymbol{\Pi}}_{nm}[k] + \rho (\V_m[k] - \bar{\X}_{nm}[k]) \\
\label{eq:admm5c}
\tilde{\boldsymbol{\Pi}}_{nm}[k+1] &=& 
\tilde{\boldsymbol{\Pi}}_{nm}[k] + \rho (\V_m[k] - \tilde{\X}_{nm}[k]) 
\end{eqnarray}
\end{subequations}

\textbf{[S2]: Primal variable update.}
\begin{subequations}
\begin{eqnarray}
\label{eq:admm6a}
\nonumber
\mathcal{P}_1[k+1] &=&
\underset{\mathcal{P}_1}{\text{arg min}} \;\;\; \mathcal{L}_{\rho} (\mathcal{P}_1, \mathcal{P}_2[k],\\
& & \;\;\;\;\;\;\;\;\;\;\;\;\;\;\;\;\;\;\;\;\;\;\;\;\;\;\; 
\mathcal{P}_3[k], \mathcal{D}[k+1]) \\
\label{eq:admm6b}
\nonumber
\mathcal{P}_2[k+1] &=&
\underset{\mathcal{P}_2}{\text{arg min}} \;\;\; \mathcal{L}_{\rho} (\mathcal{P}_1[k+1], \mathcal{P}_2, \\
& & \;\;\;\;\;\;\;\;\;\;\;\;\;\;\;\;\;\;\;\;\;\;\;\;\;\;\;
 \mathcal{P}_3[k], \mathcal{D}[k+1]) \\
\label{eq:admm6c}
\nonumber
\mathcal{P}_3[k+1] &=&
\underset{\mathcal{P}_3}{\text{arg min}} \;\;\; \mathcal{L}_{\rho} (\mathcal{P}_1[k+1], \mathcal{P}_2[k+1], \\
& & \;\;\;\;\;\;\;\;\;\;\;\;\;\;\;\;\;\;\;\;\;\;\;\;\;\;\;\;\;\;\;
 \mathcal{P}_3, \mathcal{D}[k+1]) 
\end{eqnarray}
\end{subequations}
It can be shown that the splitting variables in $\mathcal{P}_3$ turn out 
to be redundant in the final algorithm.  Letting $\bar{\boldsymbol{\Pi}}_m[k] := 
\sum_{n \in \mathcal{N}_m} \bar{\boldsymbol{\Pi}}_{nm}[k]$, it turns out that
\begin{equation}
\label{eq:admm65}
\bar{\boldsymbol{\Pi}}_m[k+1] = \bar{\boldsymbol{\Pi}}_m[k] + (\rho/2)\left(|\mathcal{N}_m| \V_m[k] - \sum_{n \in \mathcal{N}_m} \V_n[k]\right)
\end{equation}
and the dual variables $\tilde{\boldsymbol{\Pi}}_{nm}$ can be discarded. 
In addition, the per-iteration primal variable
updates per agent simplify to (see Appendix~\ref{app:decent} for derivations): 
\begin{multline}
\label{eq:admm7}
\V_m[k+1] =   \underset{\V_m \in \mathbb{R}^{N \times C}_{+}}{\text{arg min}} \;\; \Psi_{\lambda_t}(\U_m[k], \O_m[k], \V_m, \S_m^{t}, s^t) \\
 + \text{Tr}\big[ (\rho/2)|\mathcal{N}_m| \V_m^{\top}\V_m + \V_m^{\top} \big( \bar{\boldsymbol{\Pi}}_m[k+1] \\
     - (\rho/2) \{ | \mathcal{N}_m | \V_m[k] 
    + \sum_{n \in \mathcal{N}_m} \V_n[k] \} \big) \big]
\end{multline}
\begin{multline}
\label{eq:admm8}
\U_m[k+1]  \\
= \underset{\U_m \in \mathbb{R}^{N_m \times C}_{+} }{\text{arg min}} \;\; \Psi_{\lambda_t}(\U_m, \O_m[k], \V_m[k+1], \S_m^{t}, s^t)
\end{multline}
\begin{multline}
\label{eq:admm9}
\O_m[k+1]   \\
=  \underset{\O_m \in \mathbb{R}^{N_m \times C}_{+}}{\text{arg min}} \;\; \Psi_{\lambda_t}(\U_m[k+1], \O_m, \V_m[k+1], \S_m^{t}, s^t) \\
 + \text{Tr}\left( \boldsymbol{\Gamma}_m^{\top}[k+1] \O_m \right) + (\rho/2) \| \O_m - \P_m[k] \|_F^2
\end{multline}
\begin{multline}
\label{eq:admm10}
\P_m[k+1] = \underset{\P_m \in \mathbb{R}^{N_m \times C}_{+}}{\text{arg min}} (\rho/2) \| \O_m[k+1] - \P_m \|_F^2 \\
- \text{Tr}\left( \boldsymbol{\Gamma}_m^{\top}[k+1] \P_m \right) + \mu_t \| \P_m \|_1.
\end{multline}
Algorithm \ref{alg4} summarizes the
steps involved in the per-agent decentralized ADMM algorithm.
%
%
\begin{algorithm}
    \caption{Decentralized tracking algorithm per agent $m$}
\label{alg4}
\begin{algorithmic}[1]
   \STATE {\bfseries Input:}  $\{ \A_m^{t} \}_{t=1}^{T}$, $\beta$,  $C$, $\rho$
   \STATE $\U_m^0$, $\V_m^0$, $\O_m^0 = \P_m^0 = \mathbf{0}$
   \STATE $\bar{\boldsymbol{\Pi}}_m^0[0] = \mathbf{0}$, $\bar{\boldsymbol{\Gamma}}_m^0[0] = \mathbf{0}$
   \FOR{$t=1 \dots T$ }
   \STATE Set $s^t = (1-\beta^t)/(1-\beta)$
   \STATE Update $\S_m^t = \A_m^t + \beta \S_m^{t-1}$
   \STATE Update $\mu_t$ and $\lambda_t$ 
   \STATE $\V_m[0] = \V_m^{t-1}$, $\U_m[0] = \U_m^{t-1}$
   \STATE $\O_m[0] = \P_m[0]=\O_m^{t-1}$, $k=0$
   \REPEAT
   \STATE Receive $\{ \V_n[k] \}_{n \in \mathcal{N}_m}$ from neighbors of $m$
   \STATE $\boldsymbol{\Gamma}_m[k+1] = \boldsymbol{\Gamma}_m[k] + \rho (\O_m[k] - \P_m[k])$
   \STATE Compute $\bar{\boldsymbol{\Pi}}_m[k+1]$ according to \eqref{eq:admm65}
   \STATE Update $\V_m[k+1]$ via \eqref{eq:admm7}
   \STATE Update $\U_m[k+1]$ via \eqref{eq:admm8}
   \STATE Update $\O_m[k+1]$ via \eqref{eq:admm9}
   \STATE Update $\P_m[k+1]$ via \eqref{eq:admm10}
   \STATE Broadcast $\V_m[k+1]$ to single-hop neighbors
   \STATE $k = k+1$
   \UNTIL $\U_m[k], \V_m[k], \O_m[k]$ converge
   \STATE $\U_m^t = \U_m[k], \V_m^t = \V_m[k], \O_m^t = \O_m[k]$
   \ENDFOR
\end{algorithmic}
\end{algorithm}
\section{Simulations}
\label{sec:sims}
\subsection{Synthetic Data}
\label{ssec:synthdata}
\begin{figure}[ht!]
\centering
\begin{subfigure}[t]{.24\textwidth}
  \centering
  \includegraphics[width=4cm]{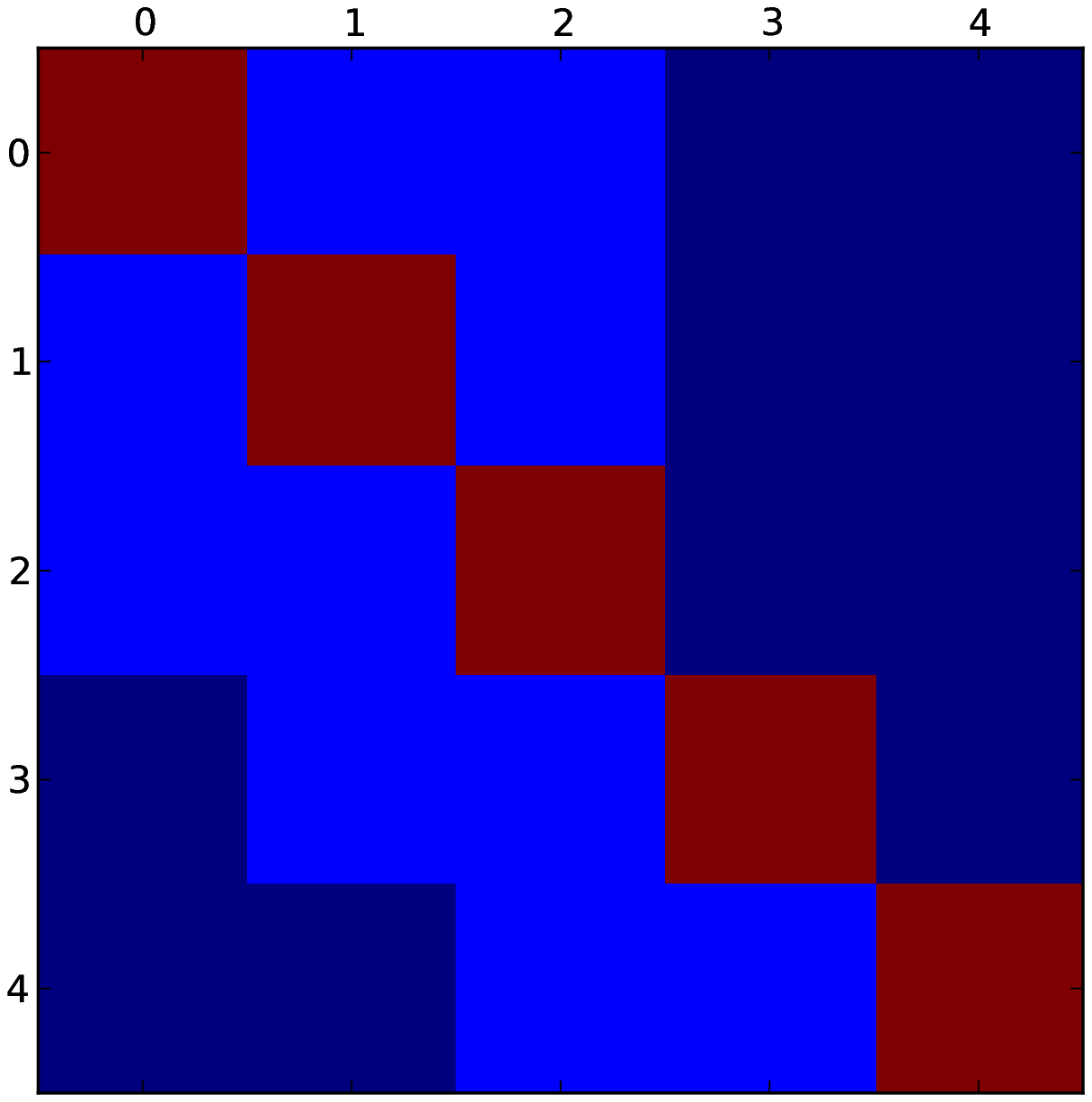}
  \caption{ } 
\end{subfigure}
\begin{subfigure}[t]{0.24\textwidth}
  \centering
 \includegraphics[width=4cm]{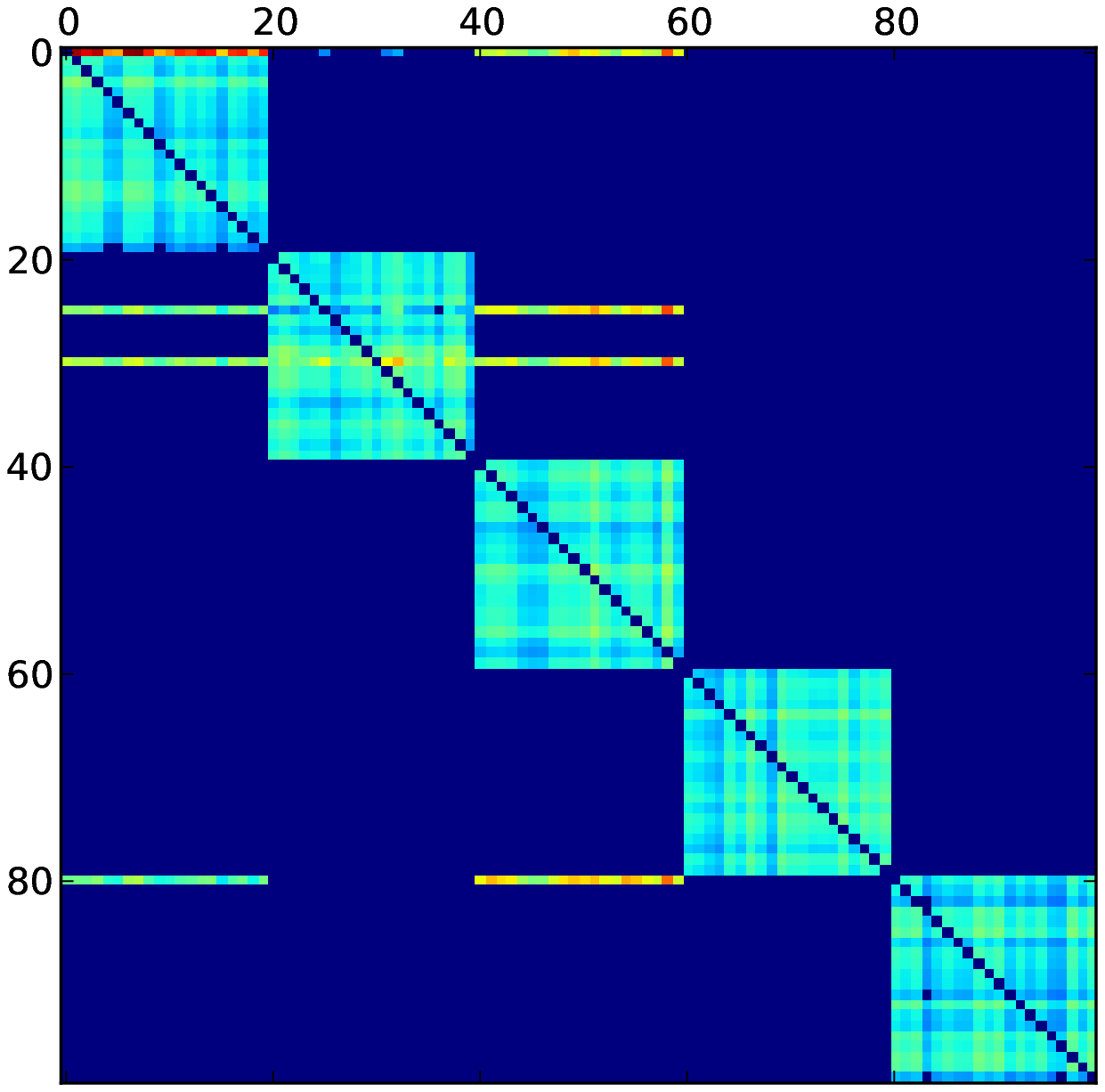}
  \caption{ } 
\end{subfigure}
\caption{ a) SBM matrix for community generation 
with parameters $\delta_{ij} \in \{0,0.1,0.8\}$, decreasing away from the diagonal; 
b) Heatmap of the initial network with anomalous nodes at rows $\{0,25,30,80\}$.}
\label{fig:hmaps}
\end{figure}

\noindent\textbf{Data generation.}
An initial synthetic network with $N = 100$, and 
$C = 5$ communities was generated using the 
stochastic blockmodel (SBM)~\cite{holland}. The
SBM parameters were set to: 
$\delta_{ii} = 0.8, i = 1,\dots,5$, and
$\delta_{ij} \in \{0, 0.1\}, i \neq j$, selected
so that the SBM matrix is
stochastic (see Figure \ref{fig:hmaps}(a)).
The initial SBM network was captured through
an adjacency matrix $\A^{\text{init}} \in \{0,1 \}^{N \times N}$.
Matrix $\A^{\text{init}}$ was then decomposed
into non-negative factors $\U^0$ and $\V^0$, that is, $\A^{\text{init}} = \U^0 (\V^0)^{\top}$,
using standard off-the-shelf NMF tools.
Anomalies were artificially induced by reconstructing a modified adjacency matrix 
$\A^0 = (\U^0 + \O^0)(\V^0)^{\top}$, where the only non-zero
rows of $\O^0$ are indexed by $\{0,25,30,80 \}$. Figure~\ref{fig:hmaps}
depicts a heatmap of $\A^0$, clearly showing anomalous nodes
as unusually highly connected. 

In order to generate slowly evolving networks, four 
piecewise constant edge-variation functions were
adopted: i) $f_1(t) = H(t)$; ii) $f_2(t) = H(t-50)$;
iii) $f_3(t) = 1 - H(t-50)$; and iv) $f_4(t) = H(t) - H(t-25) 
+ H(t-50) - H(t-75)$, where $H(t)$ denotes the unit
step function, and $t = 1,\dots,100$. The time-series
$\{ \A^t \}_{t=1}^{100}$ was generated by setting the 
edge indicators to $a_{ij}^t = a_{ij}^0 f_{\kappa}(t)$,
with $\kappa$ uniformly selected at random from $\{1,2,3,4\}$.
\begin{figure}[ht!]
\centering
\includegraphics[width=8.5cm]{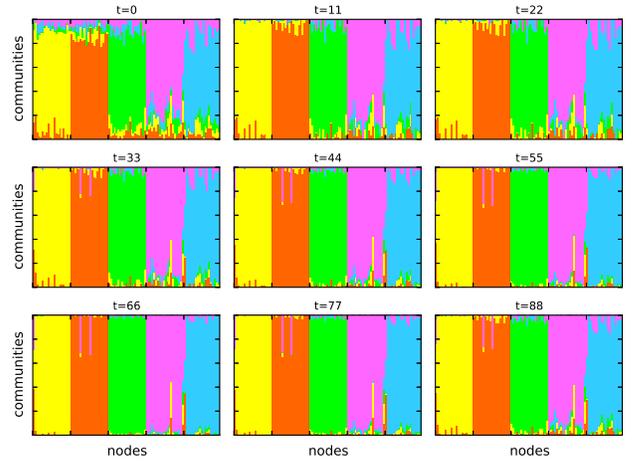}
\caption{Stacked plots depicting overlapping communities detected over a selected sample of time intervals. Horizontal axes are indexed by nodes, while vertical axes depict community affiliation
strengths that are proportional to the relative dominance of each color per node. As expected, most nodes exhibit a strong affiliation with one of the five communities.}
\label{fig:fig_synth_overlaps}
\end{figure}
\begin{figure}[ht!]
\centering
\includegraphics[width=8.5cm]{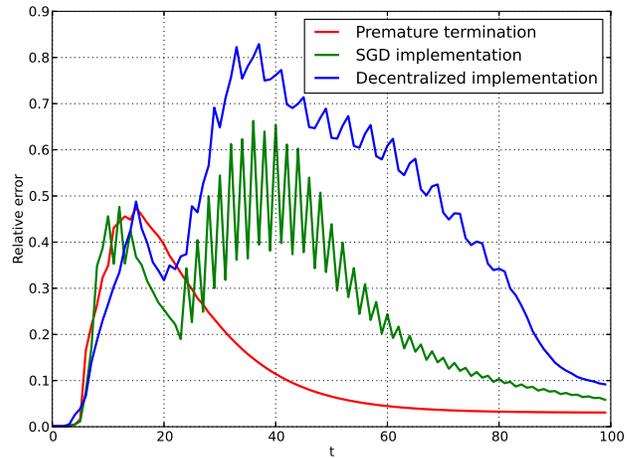}
\caption{Comparison of the relative error performance of developed 
algorithms with parameters set to $\lambda_0~=~0.05, \mu_0~=~0.1, \beta~=~0.97$.}
\label{fig:fig_compare_algs}
\end{figure}
\begin{figure*}[ht!]
\begin{minipage}[b]{.33\textwidth}
  \centering
  \includegraphics[width=5.5cm]{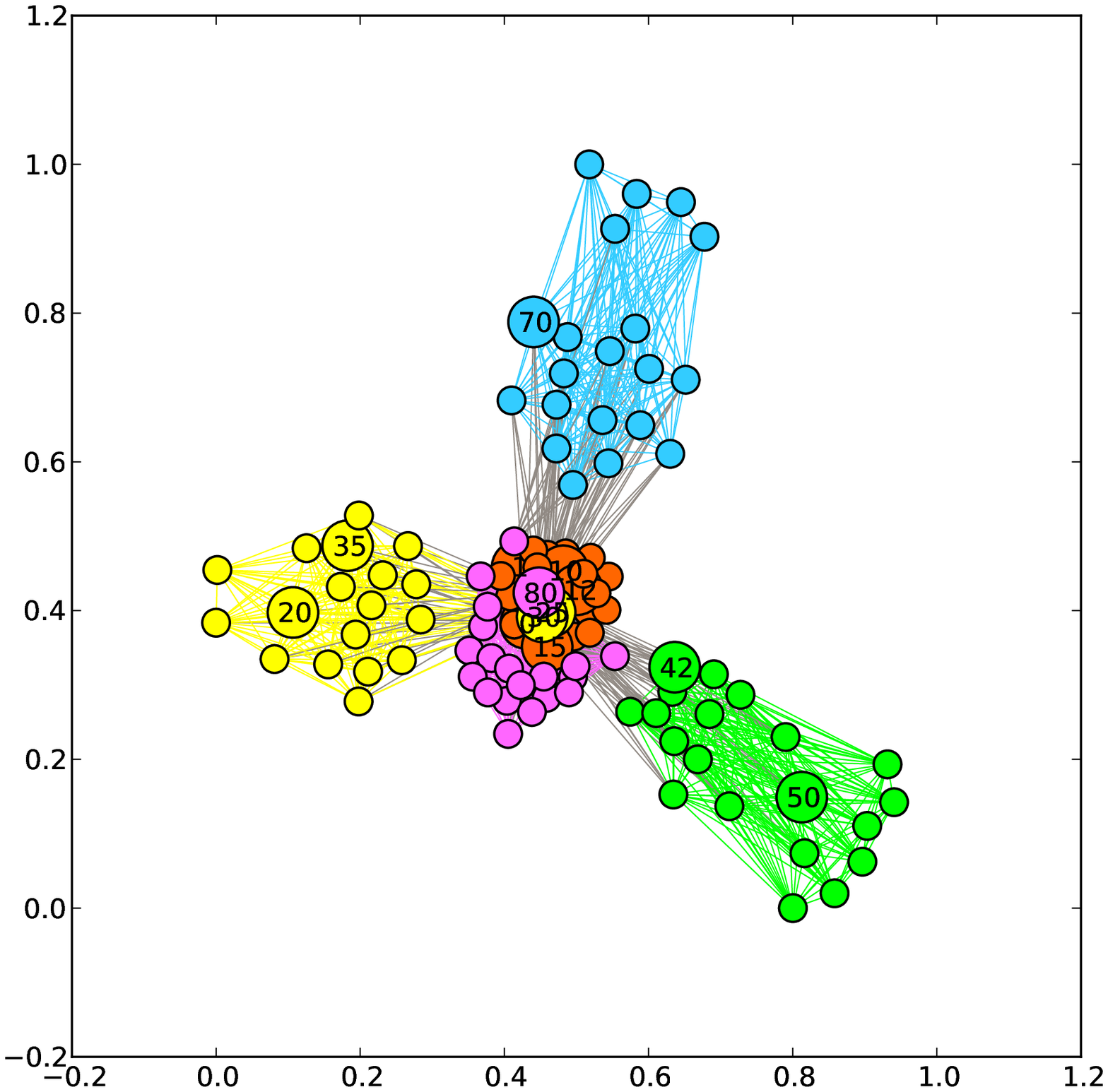}
  \centerline{(a) $ t = 10$} 
\end{minipage}
%
\begin{minipage}[b]{.33\textwidth}
  \centering
  \includegraphics[width=5.5cm]{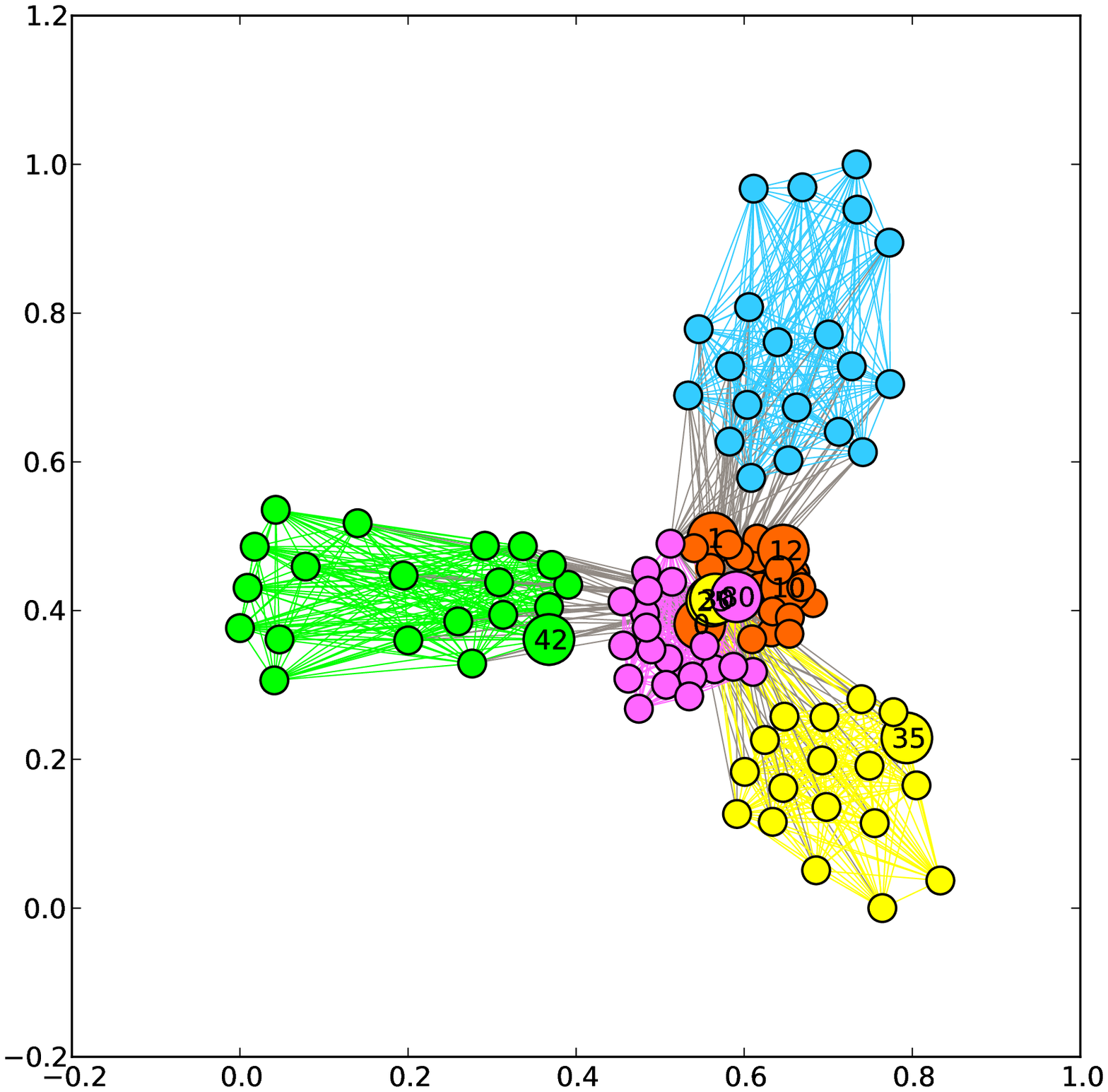}
  \centerline{(b) $ t = 40$} 
\end{minipage}
%
\begin{minipage}[b]{0.33\textwidth}
  \centering
 \includegraphics[width=5.5cm]{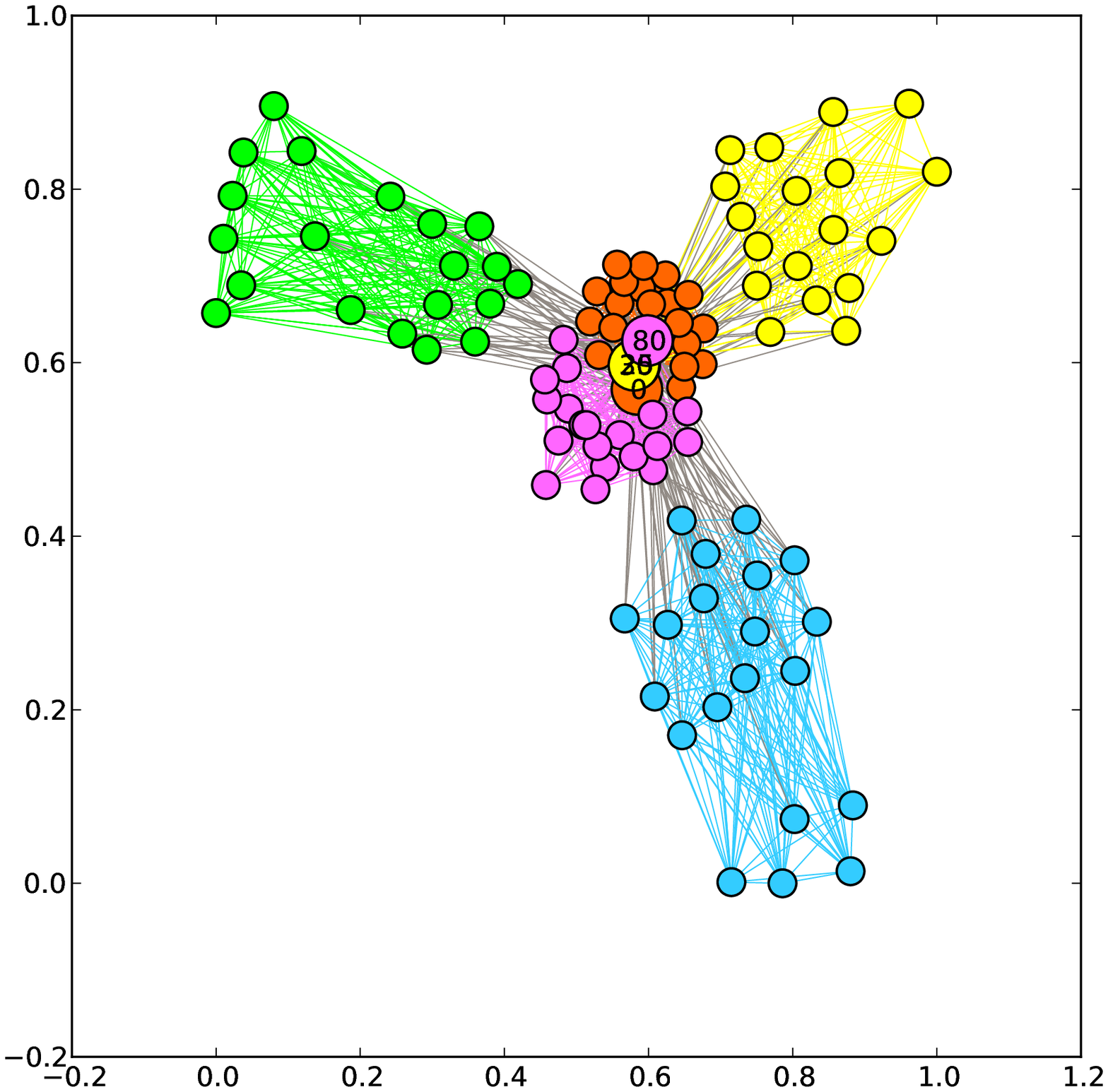}
  \centerline{(c) $ t = 70$} 
\end{minipage}
\caption{Visualization of the largest connected components 
with artificially-induced anomalies at $t=10,40,70$. Node colors
indicate detected communities, and anomalous members
are depicted by larger node sizes with labels. Although 
no anomalies were initially detected, further data acquisition 
facilitated convergence to the correct set of anomalies, i.e.,
$\{ 0,25,30,80\}$.}
\label{fig:gd_synth_anom}
\end{figure*}
\begin{figure*}[ht!]
\begin{minipage}[b]{.33\textwidth}
  \centering
  \includegraphics[width=5.5cm]{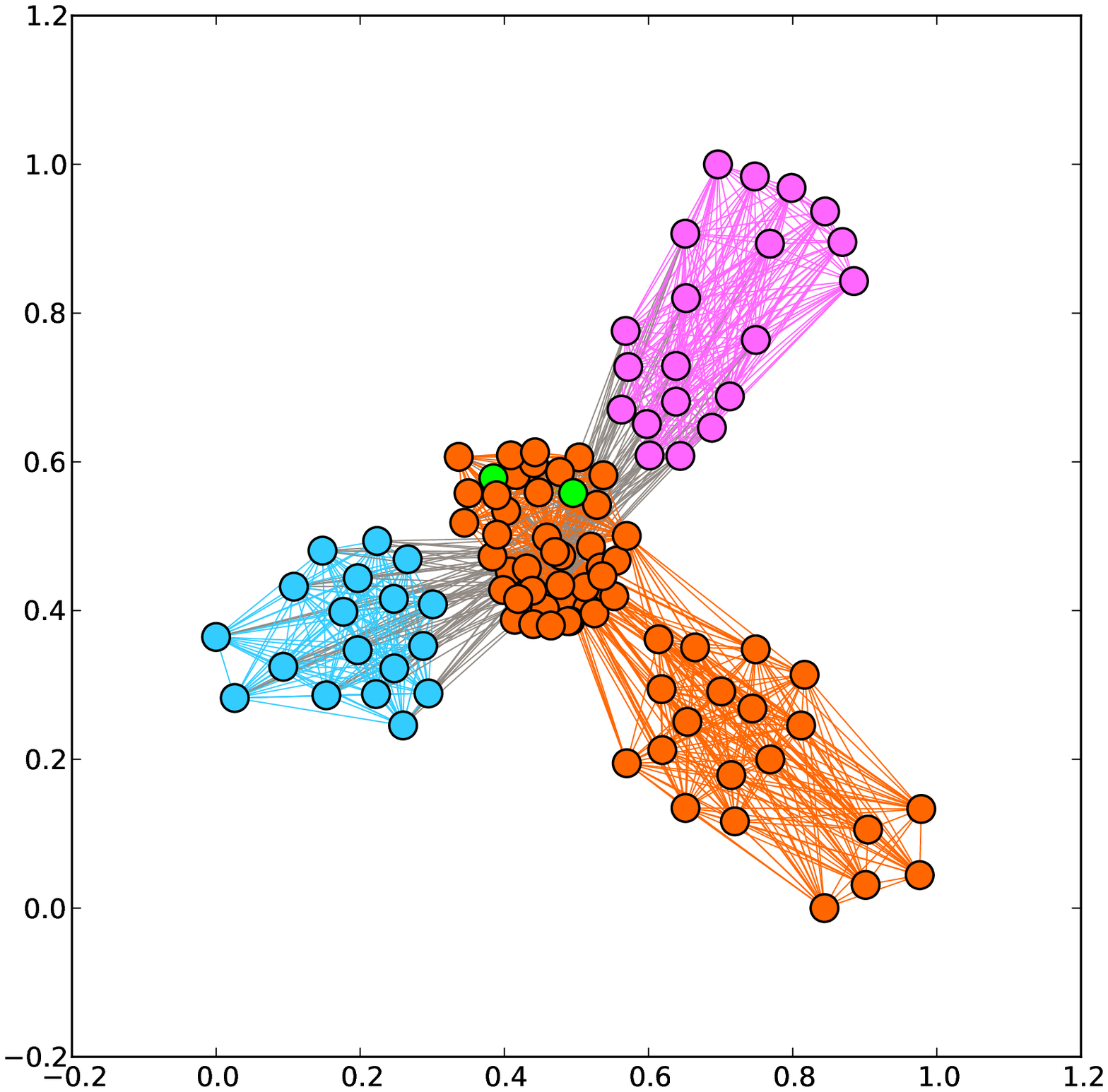}
  \centerline{(a) $ t = 10$} 
\end{minipage}
%
\begin{minipage}[b]{.33\textwidth}
  \centering
  \includegraphics[width=5.5cm]{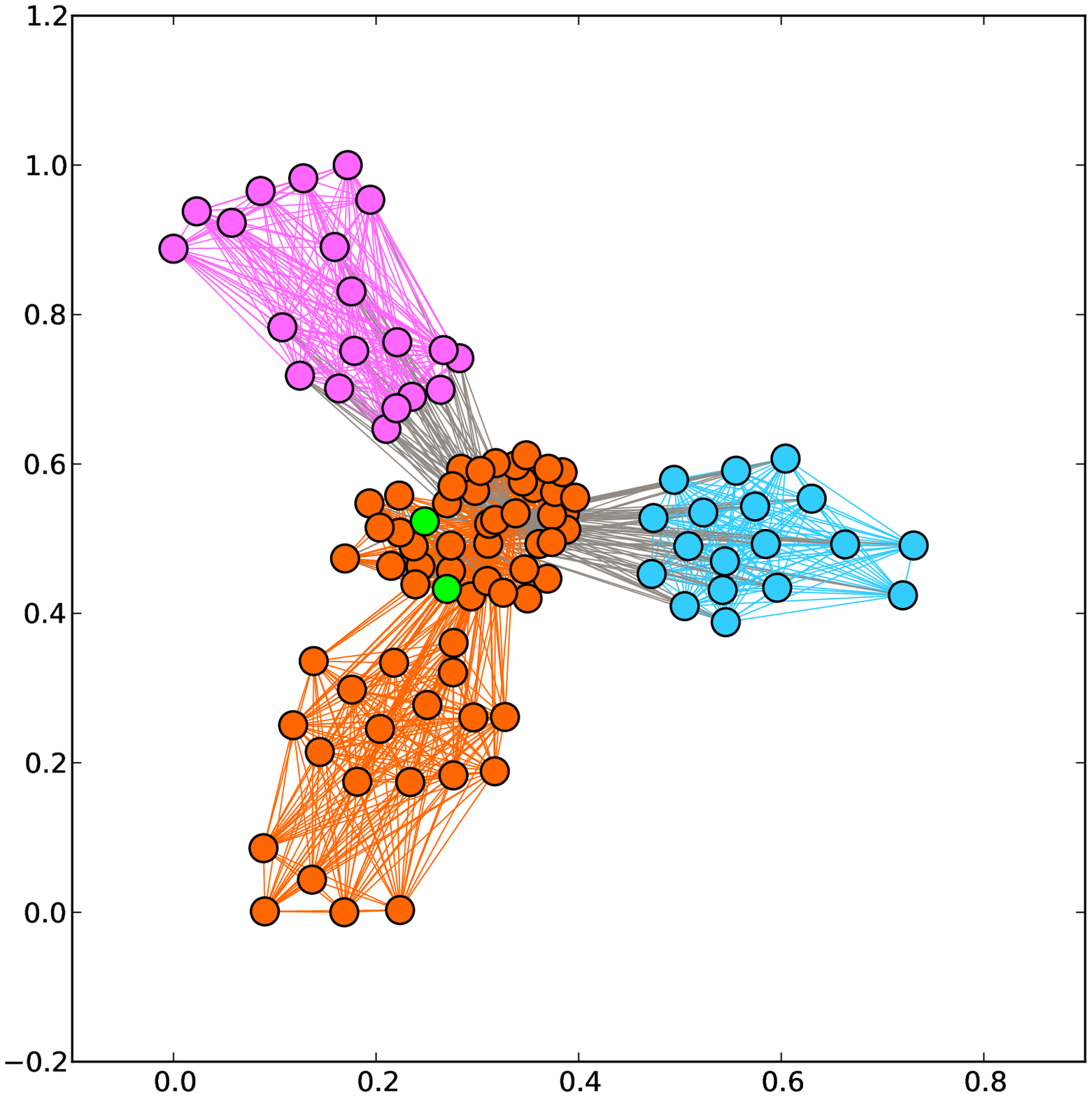}
  \centerline{(b) $ t = 40$} 
\end{minipage}
%
\begin{minipage}[b]{0.33\textwidth}
  \centering
 \includegraphics[width=5.5cm]{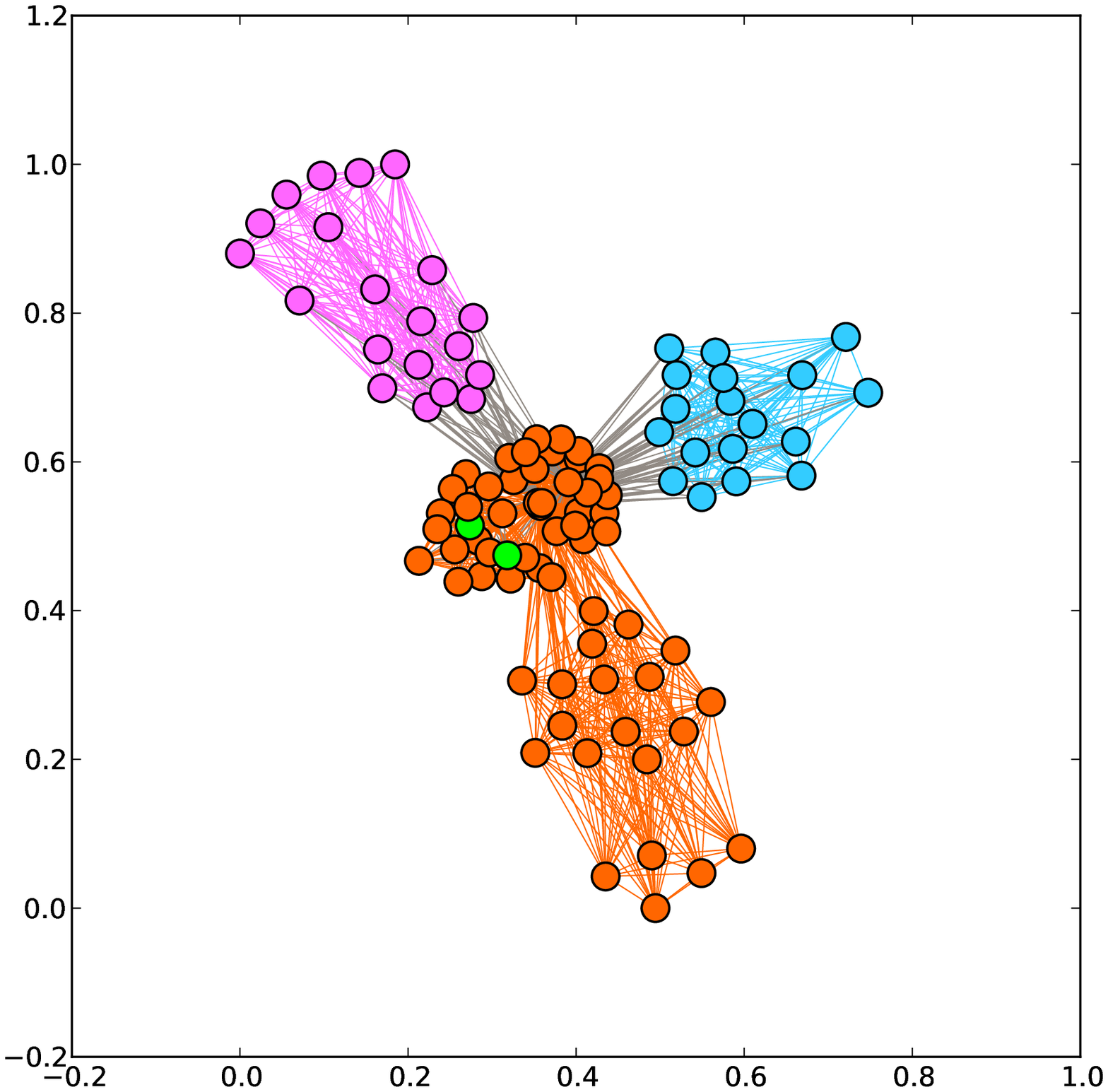}
  \centerline{(c) $ t = 70$} 
\end{minipage}
\caption{Demonstration of the distortive effect of anomalies with respect to community structure.  Plots (a)-(c) demonstrate the results of running a standard outlier-agnostic algorithm, with the inaccurate conclusion that the time-series is dominated by three communities.}
\label{fig:gd_synth_anom_agnost}
\end{figure*}
\begin{figure}[ht!]
\centering
\includegraphics[width=8.5cm]{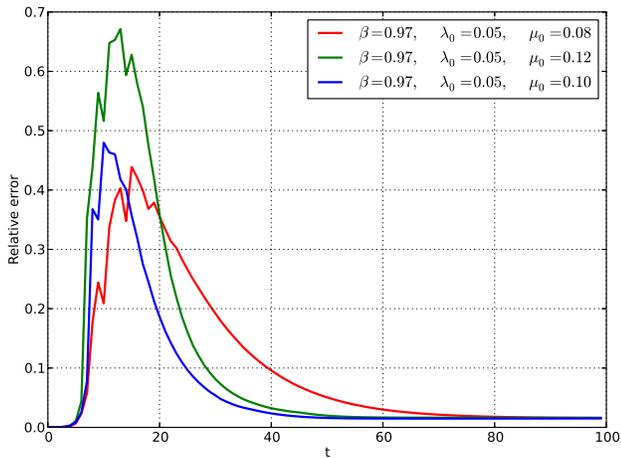}
\caption{Comparison of relative error plots resulting from running
Algorithm~\ref{alg1} and varying $\mu_0$.}
\label{fig:fig_var_mu}
\end{figure}
\begin{figure}[ht!]
\centering
\includegraphics[width=5cm]{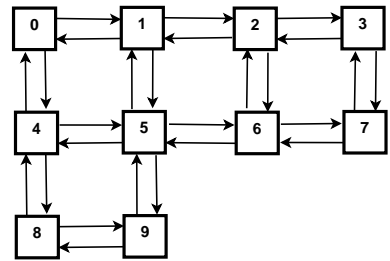}
\caption{A simple connected network of $10$ computing agents 
used for the decentralized community tracking task.}
\label{fig:agent_grid}
\end{figure}

\noindent\textbf{Numerical results.} Algorithm~\ref{alg1} was initialized by 
setting $\O$ to an all-zero
matrix, while $\U$ and $\V$ were initialized to NMF factors of $\A^0$
computed in batch. With $K=5$ and $\beta = 0.97$, Algorithm~\ref{alg1}
was run to track the constituent communities and anomalies.
Selection of $\lambda_t$ and $\mu_t$ is
admittedly challenging under dynamic settings where data are
sequentially acquired. Nevertheless, heuristics such as
increasing $\mu_t$ over time work reasonably
well when the unknowns vary slowly. It turns out that
setting $\lambda_t = 0.05$, and $\mu_t =  0.1\sqrt{t}$ yielded remarkably
good community tracking. Generally, $8-10$ inner 
iterations sufficed for convergence per time step. 
Figure~\ref{fig:fig_synth_overlaps} depicts
stacked plots of community affiliation strengths against node indices,
over a selected sample of intervals. Despite the community overlaps unveiled
by the algorithm, it is clear that all nodes generally exhibit 
strong affiliation with one of the five communities.

Hard community assignment was done by associating node $i$ with community $m$, 
where $m = \text{arg max}_k \;\; \hat{u}_{ik}$.
Figure~\ref{fig:gd_synth_anom} depicts visualizations of 
the network at $t=10$, $40$, and $70$, with node colors reflecting 
identified community membership per time interval. Nodes flagged
off as anomalies are shown with larger node sizes and explicit labels.
Initially the set of detected anomalies includes a number of 
false alarms. With acquisition of
more sequential data, the algorithm is able to iteratively prune the set,
until the ground-truth anomalies are identified (Figure~\ref{fig:gd_synth_anom} (c)). 

Further tests were conducted with Algorithms~\ref{alg2},~\ref{alg3}, and~\ref{alg4}.
For the decentralized implementation, a simple connected network of 
$10$ computing agents was adopted, as shown in Figure~\ref{fig:agent_grid}.
Figure~\ref{fig:fig_compare_algs} plots the relative error with respect to batch 
solutions, resulting from running the algorithms using the synthetic 
network time-series. The relative 
error during interval $t$ is computed as
\[
\frac{\|\hat{\U}^t - \hat{\U}^t_{\text{batch}} \|_{F}
+ \|\hat{\V}^t - \hat{\V}^t_{\text{batch}} \|_{F}
+\|\hat{\O}^t - \hat{\O}^t_{\text{batch}} \|_{F}}
{\|\hat{\U}^t_{\text{batch}}\|_F + \|\hat{\V}^t_{\text{batch}}\|_F
+\|\hat{\O}^t_{\text{batch}}\|_F}
\]
where $\hat{\U}^t_{\text{batch}}$ is the batch solution per interval 
$t$. With initial batch solutions, relative errors are initially small,
followed by dramatic increases upon acquisition of sequential data.
As more data are acquired, tracking with premature termination 
leads to faster error decay than the SGD alternative, 
presumably because it directly incorporates all 
past data by recursive aggregation. Decentralized iterations 
yield the slowest decay, because in addition to seeking convergence
to the batch solution, consensus per agent must be attained per 
time-interval. 

Selection of $\mu_t$ is critical for joint identification of anomalies,
as it controls the sparsity level inherent to $\O^t$. Although this 
is very challenging in time-varying settings, empirical 
investigation was used to guide selection of the ``best'' 
initial parameter $\mu_0$. Figure~\ref{fig:fig_var_mu}
plots the relative error (with respect to batch solutions) resulting
from running Algorithm~\ref{alg1} for several values of $\mu_0$,
with $\beta = 0.97$, and $\lambda_t = \lambda_0 = 0.05$ for all $t$.
As seen from the plots, setting $\mu_0 = 0.10$ led to the fastest
convergence to the batch solutions. 

\subsection{Real Data}
\label{ssec:realdat}
\noindent\textbf{Dataset description.} The developed algorithms
were tested on a time-series of real-world networks extracted from 
global trade flow statistics. Extracted under the \emph{Correlates of
War} project~\cite{cow_data}, the dataset captures information on 
annual bilateral trade flows (imports and exports) among countries between
$1870$ and $2009$. The network time-series were indexed by trade years,
and each node was representative of a country, while directed 
and weighted edges were indicative of the volume and direction (export/import)
of trade between countries measured in present-day U.S. dollars. 

Since trade volumes between countries can vary by orders of magnitude,
edge weights were set to logarithms of the recorded trade flows.
It is also important to note that some countries did not exist until $50$ or 
fewer years ago. As a result, network dynamics in the dataset were due to 
arrival and obsolescence of some nodes, in addition to annual changes in
edges and their weights.
Since this paper assumes that a fixed set of nodes is available, the tracking
algorithm was run for data ranging from $1949$ to $2009$ (i.e., $T = 60$), 
with $N = 170$ countries. 

The objective of this experiment was to track the 
evolution of communities within the global trade network, and to identify 
any anomalies. Note that communities in the world trade network may be interpreted as
regional trading consortia. Algorithm~\ref{alg3} was run with 
$C = 7$ communities, $\beta = 0.97$, $\alpha_u = \alpha_v = 0.002$, 
$\lambda_t = 0.5$, and $\mu_t = 1.0$, for
all $t=1,\dots,T$. Initial values $\U^0$ and $\V^0$ were obtained
by traditional NMF on $\A^0$, and $\O^0$ was set to an all-zero
$N \times C$ matrix.

\begin{table}
{\small
\begin{tabular}{|l|l|l|}
\hline
 & Country & Years as anomaly \\
\hline\hline
1 & German Federal Republic & $1954, 1955, 1956, 1966$ \\
\hline
2 & Russia & $1953, 1954, 1955, 1956$ \\
\hline
3 & Austria & $1955, 1957, 1960$ \\
\hline
4 & Pakistan & $1955, 1956, 1957$ \\
\hline
5 & Japan & $1954, 1955, 1965, 1967$ \\
\hline
6 & South Korea & $1955, 1965$ \\
\hline
7 & Yugoslavia & $1954,1956$ \\
\hline
8 & Iraq & $1967$ \\
\hline
9 & Yemen Arab Republic & $1964, 1965$ \\
\hline
10 & Kenya & $1966, 1968$ \\
\hline
\end{tabular}
}
\caption{Anomalous countries in the global trade dataset.}
\label{tab1}
\end{table}
\begin{figure}[ht!]
\centering
\includegraphics[width=8.5cm]{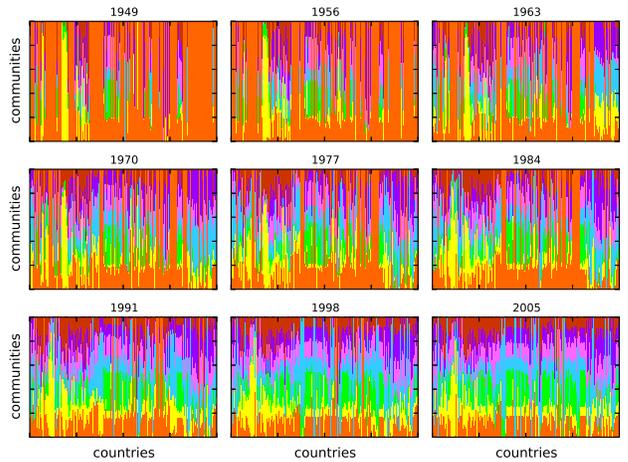}
\caption{Overlapping communities in world trade flows dataset. The bottom
row of plots suggests a growing trend of globalization, with most countries
participating in several trade communities.}
\label{fig:cow_overlap}
\end{figure}

\noindent\textbf{Numerical results.} Running Algorithm~\ref{alg3} on this dataset
revealed interesting insights about the evolution of global
trade within the last sixty years. Figure~\ref{fig:cow_overlap} depicts
stacked plots of countries and the communities they belonged to over 
a subset of years within the observation period. The horizontal axes are
indexed by countries, and each community is depicted by a specific 
color. It is clear that over the years, more countries cultivated stronger affiliations
within different communities. This observation suggests an 
increasing trend of globalization, with more countries actively 
engaging in significant trade relationships within different trade
communities. Between $1949$ and $1963$, global trade was dominated by 
one major community, while the other communities played a less significant role.
Based on historical accounts, it is likely that such trade dynamics were related
to ongoing global recovery in the years following the second world war.

\begin{figure*}[ht!]
\begin{minipage}[b]{.49\textwidth}
  \centering
  \includegraphics[width=9.5cm]{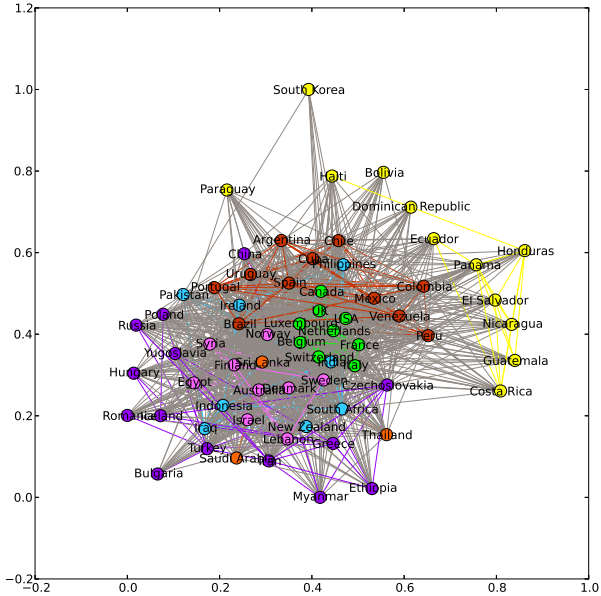}
  \centerline{(a) $1959$} 
\end{minipage}
%
\begin{minipage}[b]{0.49\textwidth}
  \centering
 \includegraphics[width=9.5cm]{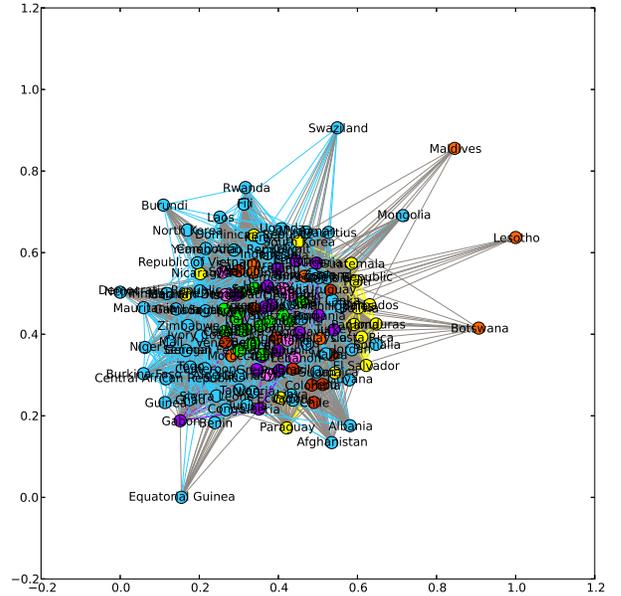}
  \centerline{(b) $1990$} 
\end{minipage}
\caption{Communities identified in the global trade network for 
the years $1959$ and $1990$.}
\label{fig:gd_real_gtn}
\end{figure*}

Figure~\ref{fig:gd_real_gtn} depicts visualizations of the global trade 
network for the years $1959$ and $1990$, with countries color-coded 
according to the community with which they are most strongly affiliated.
A core community of economic powerhouses (in green) comprises
global leaders such as the United States, United Kingdom, Canada, France etc.,
as seen from the $1959$ visualization. Interestingly, this core group
of countries remains intact as a community in $1990$. 
It turns out that geographical proximity and language play an important
role in trade relationships. This is evident from two communities
(colored maroon and yellow) which are dominated by South and Central American
nations, with Spain and Portugal as exceptions that have strong cultural and
language influences on these regions. 

Another interesting observation from the 
$1990$ visualization is the large community of developing nations (in blue).
Most of these nations only existed as colonies in $1959$, and are not depicted 
in the first drawing. However, $1990$ lies within their post-colonial
period, during which they presumably started establishing strong trading 
ties with one another. 

Finally, Table~\ref{tab1} lists countries that were flagged off by 
Algorithm~\ref{alg3} as anomalies. The table shows each of these
countries, along with a list of years during which they were identified
as anomalies. In the context of global 
trade, anomalies are expected to indicate abnormal or irregular 
trading patterns. Interestingly, the German Federal Republic, Russia,
and Japan were some of the most adversely affected countries by the 
second world war, and their trade patterns during a period of rapid economic
revival may corroborate identification as anomalies. It is also
possible that South Korea was flagged off in $1955$ and $1965$ because
of its miraculous economic growth that started in the $1950$s.

\section{Conclusion}
\label{sec:concl}
This paper put forth a novel approach for jointly tracking 
communities in time-varying network settings, and identifying
anomalous members. Leveraging advances in overlapping
community discovery, a temporal outlier-aware edge generation model was
proposed. It was assumed that the anomaly matrix is sparse,
the outlier-free, noiseless factor model is low rank, and
the network evolves slowly. Based on these assumptions,
a sparsity-promoting, rank-regularized EWLSE was advocated to 
jointly track communities and identify anomalous nodes. A first-order sequential
tracking algorithm was developed, based on alternating minimization 
and recent advances in accelerated proximal-splitting optimization.

Motivated by contemporary needs for processing big data, often
in streaming and distributed settings, a number of algorithmic 
improvements were put forth. Real-time operation was attained by
developing a fast online tracking algorithm, based on 
stochastic gradient iterations. For settings involving 
distributed acquisition and storage of network data,
a decentralized tracking algorithm that capitalizes on
the separability inherent to ADMM iterations was developed.

Simulations on synthetic SBM networks successfully
unveiled the underlying communities, and flagged off artificially-induced
anomalies. Experiments conducted on a sequence of networks
extracted from historical global trade flows between nations 
revealed interesting results concerning globalization 
of trade, and unusual trading behavior exhibited by 
certain countries during the early post-world war era.

\begin{appendices}

\section{Derivation of decentralized updates}
\label{app:decent}
Recalling the constraint $\bar{\X}_{nm} = \tilde{\X}_{nm}$, the 
per-iteration update in \eqref{eq:admm6c} becomes
\begin{multline}
\label{eqn:app1}
\bar{\X}_{nm}[k+1] = \underset{\bar{\X}_{nm}}{\text{arg min}} 
\;\; \frac{\rho}{2} \bigg\{ \| \V_m[k+1] -  \bar{\X}_{nm} \|_F^2 \\
+ \| \V_n[k+1] - \bar{\X}_{nm} \|_F^2 \bigg\} \\
  - \text{Tr} \left\lbrace \left( \bar{\boldsymbol{\Pi}}_{nm}[k+1] 
+ \tilde{\boldsymbol{\Pi}}_{nm}[k+1] \right)^{\top} \bar{\X}_{nm} \right\rbrace
\end{multline}
whose solution turns out to be
\begin{multline}
\label{eqn:app2}
\bar{\X}_{nm}[k+1] = \frac{1}{2\rho} \left( \bar{\boldsymbol{\Pi}}_{nm}[k+1] +
\tilde{\boldsymbol{\Pi}}_{nm}[k+1] \right) \\
+ \frac{1}{2} \left( \V_m[k+1] + \V_n[k+1] \right).
\end{multline}
Assuming that $\bar{\boldsymbol{\Pi}}_{nm}[0] = -\tilde{\boldsymbol{\Pi}}_{nm}[0]$,
then it can be shown by a simple inductive argument that 
$\bar{\boldsymbol{\Pi}}_{nm}[k] = -\tilde{\boldsymbol{\Pi}}_{nm}[k], 
n \in \mathcal{N}_m$~\cite{morteza_dist}. Consequently,
\begin{equation}
\label{eqn:app3}
\bar{\X}_{nm}[k] = \frac{1}{2}\left( \V_m[k] + \V_n[k] \right)
\end{equation}
and
\begin{equation}
\label{eqn:app4}
\bar{\boldsymbol{\Pi}}_{nm}[k+1] =
\bar{\boldsymbol{\Pi}}_{nm}[k] + \frac{\rho}{2} \left( \V_m[k] - \V_n[k] \right).
\end{equation}
Due to \eqref{eqn:app3}, note that $\bar{\X}_{nm}[k] = \bar{\X}_{mn}[k], n \in \mathcal{N}_m$,
and that if $\bar{\boldsymbol{\Pi}}_{nm}[0] = - \bar{\boldsymbol{\Pi}}_{mn}[0]$, then
$\bar{\boldsymbol{\Pi}}_{nm}[k] = - \bar{\boldsymbol{\Pi}}_{mn}[k], n \in 
\mathcal{N}_m$.

Focusing on the update for $\V_m$ in \eqref{eq:admm6a}, and
dropping the irrelevant terms yields
\begin{multline}
\label{eqn:app5}
\underset{\V_m}{\text{arg min}} \;\; \Psi_{\lambda_t}(\U_m[k], \O_m[k], \V_m, \S_m^t, s^t) \\
+ \text{Tr}\left\lbrace \sum\limits_{n \in \mathcal{N}_m} \left( 
\bar{\boldsymbol{\Pi}}_{nm}^{\top}[k+1] (\V_m - \bar{\X}_{nm}[k]) \right) \right\rbrace \\
+ \frac{\rho}{2} \sum\limits_{n \in \mathcal{N}_m} \| \V_m -  \bar{\X}_{nm}[k] \|_F^2.
\end{multline}
Eliminating constants in the second term, one obtains
\begin{equation}
\label{eqn:app6}
\text{Tr} \left\lbrace \sum\limits_{n \in \mathcal{N}_m} \bar{\boldsymbol{\Pi}}_{nm}^{\top}[k+1]\V_m  \right\rbrace 
= \text{Tr} \left\lbrace \V_m^{\top} \bar{\boldsymbol{\Pi}}_{m}[k+1] \right\rbrace
\end{equation}
where $ \bar{\boldsymbol{\Pi}}_{m}[k+1] := \sum\limits_{n \in \mathcal{N}_m} 
\bar{\boldsymbol{\Pi}}_{nm}[k+1]$. Using \eqref{eqn:app4},
\begin{multline}
\label{eqn:app7}
\sum\limits_{n \in \mathcal{N}_m} \bar{\boldsymbol{\Pi}}_{nm}[k+1] =
\sum\limits_{n \in \mathcal{N}_m} \bar{\boldsymbol{\Pi}}_{nm}[k] \\
+ \frac{\rho}{2}\left( |\mathcal{N}_m| - \sum\limits_{n \in \mathcal{N}_m}  \V_n[k]\right)
\end{multline}
leading to the dual variable update
\begin{equation}
\label{eqn:app8}
\bar{\boldsymbol{\Pi}}_{m}[k+1] =
\bar{\boldsymbol{\Pi}}_{m}[k]
+ \frac{\rho}{2}\left( |\mathcal{N}_m| - \sum\limits_{n \in \mathcal{N}_m}  \V_n[k]\right).
\end{equation}
Expansion of the third term in~\eqref{eqn:app5} leads to
\begin{multline}
\label{eqn:app9}
\frac{\rho}{2}\sum\limits_{n \in \mathcal{N}_m} 
\| \V_m - \bar{\X}_{nm}[k]  \|_F^2 = \\
\frac{\rho}{2} \text{Tr} \left\lbrace 
|\mathcal{N}_m| \V_m^{\top}\V_m - 2\V_m^{\top}\sum\limits_{n \in \mathcal{N}_m}
\bar{\X}_{nm}
\right\rbrace
\end{multline}
and~\eqref{eqn:app3} can be used to further expand 
$\sum_{n \in \mathcal{N}_m} \bar{\X}_{nm}$  as follows
\begin{equation}
\label{eqn:app10}
\sum\limits_{n \in \mathcal{N}_m} \bar{\X}_{nm}
= \frac{|\mathcal{N}_m|}{2} \V_m[k] + \frac{1}{2} \sum\limits_{n \in \mathcal{N}_m} \V_n[k].
\end{equation}
Upon substituting~\eqref{eqn:app10} into~\eqref{eqn:app9}, it turns out that
\begin{multline}
\label{eqn:app11}
\frac{\rho}{2}\sum\limits_{n \in \mathcal{N}_m} 
\| \V_m - \bar{\X}_{nm}[k]  \|_F^2 = 
\frac{\rho}{2} \text{Tr} \bigg\{
|\mathcal{N}_m| \V_m^{\top}\V_m \\ 
- 2\V_m^{\top} \left(
|\mathcal{N}_m| \V_m[k] + \sum\limits_{n \in \mathcal{N}_m} \V_n[k] \right)
\bigg\}
\end{multline}
and $\V_m[k+1]$ can be obtained by solving
\begin{multline}
\label{eq:app12}
\V_m[k+1] =   \underset{\V_m \in \mathbb{R}^{N \times C}_{+}}{\text{arg min}} \;\; 
\Psi_{\lambda_t}(\U_m[k], \O_m[k], \V_m, \S_m^{t}, s^t) \\
 + \text{Tr}\big[ (\rho/2)|\mathcal{N}_m| \V_m^{\top}\V_m + \V_m^{\top} \big( \bar{\boldsymbol{\Pi}}_m[k+1] \\
     - (\rho/2) \{ | \mathcal{N}_m | \V_m[k] 
    + \sum_{n \in \mathcal{N}_m} \V_n[k] \} \big) \big]
\end{multline}
whose closed-form solution is readily available. The remaining updates for 
$\U_m[k+1], \O_m[k+1],$ and $\P_m[k+1]$ follow in a straightforward manner from
the ADMM primal variable updates, and they are all available in closed form.
Note that solving for $\P_m[k+1]$ entails completion of squares, resulting
in a standard Lasso problem whose per-entry solutions are available 
in closed form using the \emph{soft thresholding} operator.
\end{appendices}

\bibliographystyle{IEEEtranS}
\bibliography{IEEEabrv,biblio}

\end{document}